\documentclass[runningheads]{CMSIM}

\usepackage{graphicx} % standard LaTeX graphics tool

\usepackage{amsmath,amssymb}                      
\usepackage{algorithm}
\usepackage[noend]{algpseudocode}
\renewcommand{\thefootnote}

\providecommand{\customgenericname}{}
\newcommand{\newcustomtheorem}[2]{  \newenvironment{#1}[1]
  {   \renewcommand\customgenericname{#2}   \renewcommand\theinnercustomgeneric{##1}   \innercustomgeneric
  }
  {\endinnercustomgeneric}
}
\newcustomtheorem{customthm}{Theorem}
\newcustomtheorem{customlemma}{Lemma}
\title*{Semi-supervised Learning Based on Distributionally Robust Optimization}

\titlerunning{\it Running Paper Title}

\author{
Jose Blanchet\inst{1}
\and
  Yang Kang\inst{2}
}

\authorrunning{\it Blanchet and Kang.}

\institute{
Management Science and Engineering, Stanford University, Stanford, CA. U.S.A.\\
(E-mail: {\tt jblanche@stanford.edu})
\and
  Department of Statistics, Columbia University, New York, NY., U.S.A.\\
  (E-mail: {\tt yang.kang@columbia.edu})
}

\begin{document}
\thispagestyle{empty}
\maketitle             
\setlength{\leftskip}{0pt}
\setlength{\headsep}{16pt}
\footnote{\begin{tabular}{p{11.2cm}r}
\small {\it $5^{th}$SMTDA Conference Proceedings, 12-15 June 2018, Chania, Creete, Greece} \\  
  \small \textcopyright {} 2018 ISAST & \includegraphics[scale=0.38]{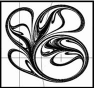}
 \end{tabular}}
\begin{abstract}
We propose a novel method for semi-supervised learning (SSL) based on
data-driven distributionally robust optimization (DRO) using optimal transport
metrics. Our proposed method enhances generalization error by using the
non-labeled data to restrict the support of the worst case distribution in our
DRO formulation. We enable the implementation of our DRO formulation by
proposing a stochastic gradient descent algorithm which allows to easily
implement the training procedure. We demonstrate that our Semi-supervised DRO
method is able to improve the generalization error over natural supervised
procedures and state-of-the-art SSL estimators. Finally, we include a
discussion on the large sample behavior of the optimal uncertainty region in
the DRO formulation. Our discussion exposes important aspects such as the role
of dimension reduction in SSL.
\end{abstract}

\section{Introduction}
\vspace{-3mm}
We propose a novel method for semi-supervised learning (SSL) based on
data-driven distributionally robust optimization (DRO) using an optimal
transport metric -- also known as earth-moving distance (see
\cite{rubner2000earth}).

\begin{figure}[pth]
\centering
\includegraphics[width=10cm,height = 3.5cm]{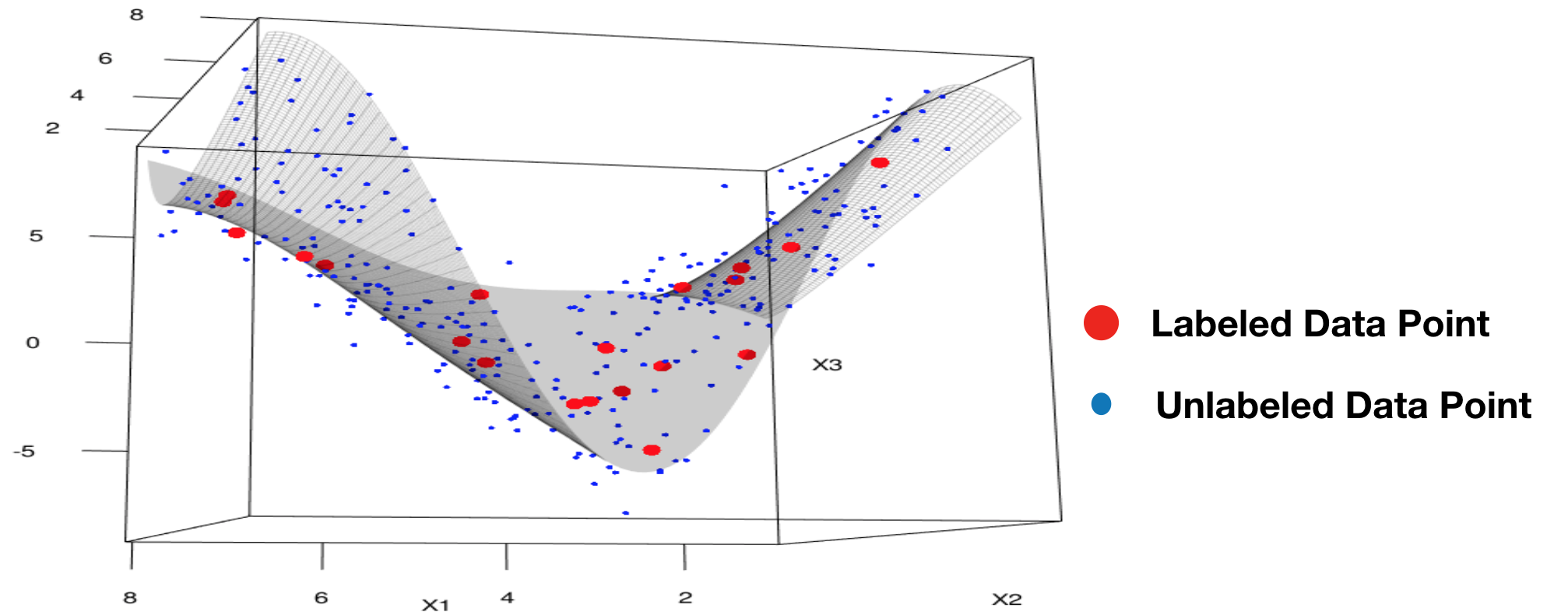}
\caption{Idealization of the way in which the unlabeled predictive variables
provide a proxy for an underlying lower dimensional manifold. Large red dots
represent labeled instances and small blue dots represent unlabeled
instances.}%
\label{Fig_DRO_intui1}%
\vspace{-5mm}
\end{figure}

Our approach enhances generalization error by using the unlabeled data to
restrict the support of the models which lie in the region of distributional
uncertainty. The intuition is that our mechanism for fitting the underlying
model is automatically tuned to generalize beyond the training set, but only
over potential instances which are relevant. The expectation is that
predictive variables often lie in lower dimensional manifolds embedded in the
underlying ambient space; thus, the shape of this manifold is informed by the
unlabeled data set (see Figure \ref{Fig_DRO_intui1} for an illustration of
this intuition).\ 

%\begin{figure}[pth]
%\begin{center}
%\includegraphics[width=\textwidth]{General_plot}
%\end{center}
%\caption{(a): Idealization of the way in which the unlabeled
%		predictive variables provide a proxy for an underlying lower dimensional
%		manifold. Large red dots represent labeled instances and small blue dots
%		represent unlabeled instances. (b)Pictorial representation of the role that the support
%		constraint plays in the SSL-DRO approach and how its presence
%		enhances the out-of-sample performance.}%
%\label{Fig_DRO_intui0}%
%\end{figure}

To enable the implementation of the DRO formulation we propose a stochastic
gradient descent (SGD) algorithm which allows to implement the training
procedure at ease. Our SGD construction includes a procedure of independent
interest which, we believe, can be used in more general stochastic
optimization problems.

We focus our discussion on semi-supervised classification but the modeling and
computational approach that we propose can be applied more broadly as we shall
illustrate in Section \ref{Sec-Error-Improve}.

We now explain briefly the formulation of our learning procedure. Suppose that
the training set is given by $\mathcal{D}_{n}=\{\left(  Y_{i},X_{i}\right)
\}_{i=1}^{n}$, where $Y_{i}\in\{-1,1\}$ is the label of the $i$-th observation
and we assume that the predictive variable, $X_{i}$, takes values in
$\mathbb{R}^{d}$. We use $n$ to denote the number of labeled data points.

In addition, we consider a set of unlabeled observations, $\left\{
X_{i}\right\}  _{i=n+1}^{N}$. We build the set $\mathcal{E}_{N-n}=\{\left(
1,X_{i}\right)  \}_{i=n+1}^{N}\cup\{\left(  -1,X_{i}\right)  \}_{i=n+1}^{N}$.
That is, we replicate each unlabeled data point twice, recognizing that the
missing label could be any of the two available alternatives. We assume that
the data must be labeled either -1 or 1.

We then construct the set $\mathcal{X}_{N}=\mathcal{D}_{n}\cup\mathcal{E}%
_{N-n}$ which, in simple words, is obtained by just combining both the labeled
data and the unlabeled data with all the possible labels that can be assigned.
The cardinality of $\mathcal{X}_{N}$, denoted as $\left\vert \mathcal{X}%
_{N}\right\vert $, is equal to $2\left(  N-n\right)  +n$ (for simplicity we
assume that all of the data points and the unlabeled observations are distinct).

Let us define $\mathcal{P}\left(  \mathcal{X}_{N}\right)  $ to be the space of
probability measures whose support is contained in $\mathcal{X}_{N}$. We use
$P_{n}$ to denote the empirical measure supported on the set $\mathcal{D}_{n}%
$, so $P_{n}\in\mathcal{P}\left(  \mathcal{X}_{N}\right)  $. In addition, we
write $E_{P}\left(  \cdot\right)  $ to denote the expectation associated with
a given probability measure $P$.

Let us assume that we are interested in fitting a classification model by
minimizing a given expected loss function $l\left(  X,Y,\beta\right)  $, where
$\beta$ is a parameter which uniquely characterizes the underlying model. We
shall assume that $l\left(  X,Y,\cdot\right)  $ is a convex function for each
fixed $\left(  X,Y\right)  $. \ The empirical risk associated to the parameter
$\beta$ is%
\[
E_{P_{n}}\left(  l\left(  X,Y,\beta\right)  \right)  =\frac{1}{n}\sum
_{i=1}^{n}l\left(  X_{i},Y_{i},\beta\right)  .
\]
In this paper, we propose to estimate $\beta$ by solving the DRO problem%
\begin{equation}
\min_{\beta}\max_{P\in\mathcal{P}\left(  \mathcal{X}_{N}\right)  :D_{c}\left(
P,P_{n}\right)  \leq\delta^{\ast}}E_{P}[l\left(  X,Y,\beta\right)  ],
\label{DRO_1}%
\end{equation}
where $D_{c}\left(  \cdot\right)  $ is a suitably defined discrepancy between
$P_{n}$ and any probability measure $P\in\mathcal{P}\left(  \mathcal{X}%
_{N}\right)  $ which is within a certain tolerance measured by $\delta^{\ast}$.

So, intuitively, (\ref{DRO_1}) represents the value of a game in which the
outer player (we) will choose $\beta$ and the adversary player (nature) will
rearrange the support and the mass of $P_{n}$ within a budget measured by
$\delta^{\ast}$. We then wish to minimize the expected risk regardless of the
way in which the adversary might corrupt (within the prescribed budget) the
existing evidence. In formulation (\ref{DRO_1}), the adversary is crucial to
ensure that we endow our mechanism for selecting $\beta$ with the ability to
cope with the risk impact of out-of-sample (i.e. out of the training set)
scenarios. We denote the formulation in \eqref{DRO_1} as semi-supervised
distributionally robust optimization (SSL-DRO).

The criterion that we use to define $D_{c}\left(  \cdot\right)  $ is based on
the theory of optimal transport and it is closely related to the concept of
Wasserstein distance, see Section \ref{Sec_sem_DRO}. The choice of
$D_{c}\left(  \cdot\right)  $ is motivated by recent results which show that
popular estimators such as regularized logistic regression, Support Vector
Machines (SVMs) and square-root Lasso (SR-Lasso) admit a DRO representation
\textit{exactly equal to} (\ref{DRO_1}) in which the support $\mathcal{X}_{N}$
is replaced by $\mathbb{R}^{d+1}$ (see \cite{blanchet2016robust} and also
equation \eqref{Eqn-Reg_LR-Rep} in this paper.)

In view of these representation results for supervised learning algorithms,
the inclusion of $\mathcal{X}_{N}$ in our DRO formulation (\ref{DRO_1})
provides a natural SSL approach in the context of classification and
regression. The goal of this paper is to enable the use of the
distributionally robust training framework (\ref{DRO_1}) as a SSL technique.
We will show that estimating $\beta$ via (\ref{DRO_1}) may result in a
significant improvement in generalization relative to natural supervised
learning counterparts (such as regularized logistic regression and SR-Lasso).
The potential improvement is illustrated in Section \ref{Sec-Error-Improve}.
Moreover, we show via numerical experiments in Section \ref{Sec_Numerical},
that our method is able to improve upon state-of-the-art SSL algorithms.

As a contribution of independent interest, we construct a stochastic gradient
descent algorithm to approximate the optimal selection, $\beta_{N}^{\ast}$,
minimizing (\ref{DRO_1}).

An important parameter when applying (\ref{DRO_1}) is the size of the
uncertainty region, which is parameterized by $\delta^{\ast}$. We apply
cross-validation to calibrate $\delta^{\ast}$, but we also discuss the
non-parametric behavior of an optimal selection of $\delta^{\ast}$ (according
to a suitably defined optimality criterion explained in Section
\ref{Sec_del_star}) as $n,N\rightarrow\infty$.

In Section \ref{Section_Alternative}, we provide a broad overview of
alternative procedures in the SSL literature, including recent approaches
which are related to robust optimization. A key role in our formulation is
played by $\delta^{\ast}$, which can be seen as a regularization parameter.
This identification is highlighted in the form of (\ref{DRO_1}) and the
DRO\ representation of regularized logistic regression which we recall in
\eqref{Eqn-Reg_LR-Rep}. The optimal choice of $\delta^{\ast}$ ensures
statistical consistency as $n,N\rightarrow\infty$.

Similar robust optimization formulations to (\ref{DRO_1}) for machine learning
have been investigated in the literature recently. For example, connections
between robust optimization and machine learning procedures such as Lasso and
SVMs have been studied in the literature, see \cite{xu2009robust}. In contrast
to this literature, the use of distributionally robust uncertainty allows to
discuss the optimal size of the uncertainty region as the sample size
increases (as we shall explain in Section \ref{Sec_del_star}). The work of
\cite{shafieezadeh2015distributionally} is among the first to study
DRO\ representations based on optimal transport, they do not study the
implications of these types of DRO\ formulations in SSL as we do here.

We close this Introduction with a few important notes. First, our SSL-DRO is
not a robustifying procedure for a given SSL algorithm. Instead, our
contribution is in showing how to use unlabeled information on top of DRO to
enhance traditional supervised learning methods. In addition, our SSL-DRO
formulation, as stated in \eqref{DRO_1} , is not restricted to logistic
regression, instead DRO counterpart could be formulated for general supervised
learning methods with various choice of loss function.

%The rest of the paper is structured as follows. We will quickly review the
%alternative related state-of-the-art SSL algorithms. In Section
%\ref{Sec_sem_DRO} we discuss the elements of our DRO\ formulation, including
%the definition of optimal transport metric and the implementation of a
%stochastic gradient descent algorithm for the solution of (\ref{DRO_1}). In
%Section \ref{Sec-Error-Improve} we explore the improvement in out-of-sample
%performance of our method relative to regularized logistic regression. In
%Section \ref{Sec_Numerical}, we compare our procedure against alternative SSL
%estimators, both in the context of some binary classification real data sets.
%In Section \ref{Sec_del_star}, we explore the behavior of the optimal
%uncertainty size $\delta^{\ast}$ as the sample size increases, especially we
%discuss certain asymptotic results on how to pick up the distributional
%uncertainty size optimally with asymptotic consistency. Section \ref{Sec-Con}
%contains final considerations and further discussions. In Appendix
%\ref{Sec-Appendix-Tech}, we provide more
%technical details for the asymptotic results stated in Section
%\ref{Sec_del_star}.
\vspace{-3mm}
\section{Alternative Semi-Supervised Learning
\vspace{-3mm}
Procedures\label{Section_Alternative}}

We shall briefly discuss alternative procedures which are known in the SSL
literature, which are quite substantial. We refer the reader to the excellent
survey of \cite{zhu2005semi} for a general overview of the area. Our goal here
is to expose the similarities and connections between our approach and some of
the methods that have been adopted in the community.

For example, broadly speaking graph-based methods \cite{blum2001learning} and
\cite{chapelle2009semi} attempt to construct a graph which represents a sketch
of a lower dimensional manifold in which the predictive variables lie. Once
the graph is constructed, a regularization procedure is performed, which seeks
to enhance generalization error along the manifold while ensuring continuity
in the prediction regarding an intrinsic metric. Our approach bypasses the
construction of the graph, which we see as a significant advantage of our
procedure. However, we believe that the construction of the graph can be used
to inform the choice of cost function $c\left(  \cdot\right)  $ which should
reflect high transportation costs for moving mass away from the manifold
sketched by the graph.

%Another class of popular SSL
%algorithms are so-called Transductive Support Vector Machines (TSVM),
%see\cite{vapnik1998statistical,joachims1999transductive}
%for a general overview. It follows, see Section 3.2 in
%\cite{blanchet2016robust}, that SVMs
%can be represented by means of a DRO representation such as (\ref{DRO_1})
%without the constraint that $P\in\mathcal{P}\left(  \mathcal{X}_{N}\right)  $.
%Our approach provides another natural extension of semi-supervised SVMs (other
%than TSVM) by simply adding that $P\in\mathcal{P}\left(  \mathcal{X}%
%_{N}\right)  $ as we do in (\ref{DRO_1}).
%Further study
%between TSVMs and our approach are interesting topics for further study. One
%advantage that is apparent in our approach is that it does not lead to the
%types of non-convex features that arise in the training of TSVMs, see for
%example the discussion in p. 15 of
%\cite{zhu2005semi}. So, the stochastic
%gradient descent procedure explained in (\ref{SGD}) can be directly applied in
%our version of semi-supervised SVMs. We plan to report on this line of
%research in future work.

Some recent SSL estimators are based on robust optimization, such as the work
of \cite{balsubramani2015scalable}. The difference between data-driven DRO and
robust optimization is that the inner maximization in (\ref{DRO_1}) for robust
optimization is not over probability models which are variations of the
empirical distribution. Instead, in robust optimization, one attempts to
minimize the risk of the worst case performance of potential outcomes inside a
given uncertainty set. 

In \cite{balsubramani2015scalable}, the robust uncertainty set is defined in
terms of constraints obtained from the testing set. The problem with the
approach in \cite{balsubramani2015scalable} is that there is no clear
mechanism which informs an optimal size of the uncertainty set (which in our
case is parameterized by $\delta^{\ast}$). In fact, in the last paragraph of
Section 2.3, \cite{balsubramani2015scalable} point out that the size of the
uncertainty could have a significant detrimental impact in practical
performance. 

We conclude with a short discussion on the work of \cite{loog2016contrastive},
which is related to our approach. In the context of linear discriminant
analysis, \cite{loog2016contrastive} also proposes a distributionally robust
optimization estimator, although completely different from the one we propose
here. More importantly, we provide a way (both in theory and practice) to
study the optimal size of the distributional uncertainty (i.e. $\delta^{\ast}%
$), which allows us to achieve asymptotic consistency of our estimator.
\vspace{-3mm}
\section{Semi-supervised Learning based on DRO\label{Sec_sem_DRO}}
\vspace{-3mm}
This section is divided into two parts. First, we provide the elements of our
DRO formulation. Then we will explain how to solve the SSL-DRO problem, i.e.
find optimal $\beta$ in (\ref{DRO_1}).
\vspace{-5mm}
\subsection{Defining the optimal transport discrepancy:}
\vspace{-2mm}
Assume that the cost function $c:\mathbb{R}^{d+1}\times\mathbb{R}%
^{d+1}\rightarrow\lbrack0,\infty]$ is lower semi-continuous. As mentioned in
the Introduction, we also assume that $c(u,v)=0$ if and only if $u=v$.

Now, given two distributions $P$ and $Q$, with supports $\mathcal{S}%
_{P}\subseteq\mathcal{X}_{N}$ and $\mathcal{S}_{Q}\subseteq\mathcal{X}_{N}$,
respectively, we define the optimal transport discrepancy, $D_{c}$, via%
\begin{align}
D_{c}\left(  P,Q\right)  =\inf\{E_{\pi}\left[  c\left(  U,V\right)  \right]
:\pi\in\mathcal{P}\left(  \mathcal{S}_{P}\times\mathcal{S}_{Q}\right)  ,\text{
}\pi_{U}=P,\text{ }\pi_{V}=Q\}, \label{Discrepancy_Def}%
\end{align}
where $\mathcal{P}\left(  \mathcal{S}_{P}\times\mathcal{S}_{Q}\right)  $ is
the set of probability distributions $\pi$ supported on $\mathcal{S}_{P}%
\times\mathcal{S}_{Q}$, and $\pi_{U}$ and $\pi_{V}$ denote the marginals of
$U$ and $V$ under $\pi$, respectively.

If, in addition, $c\left(  \cdot\right)  $ is symmetric (i.e. $c\left(
u,v\right)  =c\left(  v,u\right)  $), and there exists $\varrho\geq1$ such
that $c^{1/\varrho}\left(  u,w\right)  \leq c^{1/\varrho}\left(  u,v\right)
+c^{1/\varrho}\left(  v,w\right)  $ (i.e. $c^{1/\varrho}\left(  \cdot\right)
$ satisfies the triangle inequality), it can be easily verified (see
\cite{villani2008optimal}) that $D_{c}^{1/\varrho}\left(  P,Q\right)  $ is a
metric. For example, if $c\left(  u,v\right)  =\left\Vert u-v\right\Vert
_{q}^{\varrho}$ for $q\geq1$ (where $\left\Vert u-v\right\Vert _{q}$ denotes
the $l_{q}$ norm in $\mathbb{R}^{d+1}$) then $D_{c}\left(  \cdot\right)  $ is
known as the Wasserstein distance of order $\varrho$.

Observe that (\ref{Discrepancy_Def}) is obtained by solving a linear
programming problem. For example, suppose that $Q=P_{n}$, and let
$P\in\mathcal{P}\left(  \mathcal{X}_{N}\right)  $ then, using $U=\left(
X,Y\right)  $, we have that $D_{c}\left(  P,P_{n}\right)  $ is obtained by
computing
%\begin{align}\label{LP}
%	 \min_{\pi} & \sum_{u\in\mathcal{X}_{N}} \sum_{v\in\mathcal{D}_{n}}c\left(  u,v\right) \pi\left(  u,v\right) \\
%	\mbox{ s.t.} &  \sum_{u\in\mathcal{X}_{N}}\pi\left(  u,v\right)  =\frac{1}{n}\text{
%	}\forall\text{ }v\in\mathcal{D}_{n},\nonumber\\
%	& sum_{v\in\mathcal{D}_{N}}\pi\left(  u,v\right)  =P\left(  \left\{
%	u\right\}  \right)  \text{ }\forall\text{ }u\in\mathcal{X}_{N}, \nonumber\\
%	&\pi\left(  u,v\right)  \geq0\text{ }\forall\text{ }\left(
%	u,v\right)  \in\mathcal{X}_{N}\times\mathcal{D}_{n}\nonumber
%\end{align}
%

\begin{align}
\label{LP}\min_{\pi} \big\{ & \sum_{u\in\mathcal{X}_{N}} \sum_{v\in
\mathcal{D}_{n}}c\left(  u,v\right)    \pi\left(  u,v\right)  :\mbox{ s.t.}
\sum_{u\in\mathcal{X}_{N}}\pi\left(  u,v\right)  =\frac{1}{n}\text{ }%
\forall\text{ }v\in\mathcal{D}_{n},\\
&  \sum_{v\in\mathcal{D}_{N}}\pi\left(  u,v\right)  =P\left(  \left\{
u\right\}  \right)  \text{ }\forall\text{ }u\in\mathcal{X}_{N}, \pi\left(
u,v\right)  \geq0\text{ }\forall\text{ }\left(  u,v\right)  \in\mathcal{X}%
_{N}\times\mathcal{D}_{n} \big\}\nonumber
\end{align}
We shall discuss, for instance, how the choice of $c\left(  \cdot\right)  $ in
formulations such as (\ref{DRO_1}) can be used to recover popular machine
learning algorithms.
\vspace{-5mm}
\subsection{Solving the SSL-DRO formulation:}
\vspace{-2mm}
A direct approach to solve (\ref{DRO_1}) would involve alternating between
minimization over $\beta$, which can be performed by, for example, stochastic
gradient descent and maximization which is performed by solving a linear
program similar to (\ref{LP}). Unfortunately, the large scale of the linear
programming problem, which has $O(N)$ variables and $O(n)$ constraints, makes
this direct approach rather difficult to apply in practice. \newline
So, our goal here is to develop a direct stochastic gradient descent approach
which can be used to approximate the solution to (\ref{DRO_1}). 
\newline First, it is useful to apply linear programming duality to simplify
(\ref{DRO_1}). Note that, given $\beta$, the inner maximization in
(\ref{DRO_1}) is simply
%\begin{align}\label{LP_2}
%	\max_{\pi} & \sum_{u\in\mathcal{X}_{N}}\sum_{v\in\mathcal{D}_{N}%
%	}l\left(  u,\beta\right)  \pi\left(  u,v\right)\\
%	s.t. & \sum_{u\in\mathcal{X}_{N}}\pi\left(  u,v\right)
%	=\frac{1}{n}\text{ }\forall\text{ }v\in\mathcal{D}_{n} \nonumber\\
%	&\sum_{u\in\mathcal{X}_{N}}\sum_{v\in\mathcal{D}_{n}%
%	}c\left(  u,v\right)  \pi\left(  u,v\right)  \leq\delta,\nonumber\\
%	& \pi\left(  u,v\right)  \geq0\text{ }\forall\text{ }\left(
%	u,v\right)  \in\mathcal{X}_{N}\times\mathcal{D}_{n}. \nonumber
%\end{align}
%
\begin{align}
\label{LP_2}\max_{\pi}\big\{& \sum_{u\in\mathcal{X}_{N}}\sum_{v\in
\mathcal{D}_{N}}l\left(  u,\beta\right)     \pi\left(  u,v\right)  :
\mbox{ s.t.} \sum_{u\in\mathcal{X}_{N}}\pi\left(  u,v\right)  =\frac{1}%
{n}\text{ }\forall\text{ }v\in\mathcal{D}_{n}\\
&  \sum_{u\in\mathcal{X}_{N}}\sum_{v\in\mathcal{D}_{n}}c\left(  u,v\right)
\pi\left(  u,v\right)  \leq\delta, \pi\left(  u,v\right)  \geq0\text{ }%
\forall\text{ }\left(  u,v\right)  \in\mathcal{X}_{N}\times\mathcal{D}_{n}
\big\}.\nonumber
\end{align}
Of course, the feasible region in this linear program is always non-empty
because the probability distribution $\pi\left(  u,v\right)  =I\left(
u=v\right)  I\left(  v\in\mathcal{D}_{n}\right)  /n$ is a feasible choice.
Also, the feasible region is clearly compact, so the dual problem is always
feasible and by strong duality its optimal value coincides with that of the
primal problem, see \cite{bertsimas2011theory,bertsimas2013data,blanchet2016robust}. 
The dual problem associated to
(\ref{LP_2}) is given by%
\begin{align}
\min\big\{ \sum_{v\in\mathcal{D}_{N}}\gamma\left(  v\right)  /n+\lambda
\delta\text{ s.t. }  &  \gamma\left(  v\right)  \in\mathbb{R}\text{ }%
\forall\text{ }v\in\mathcal{D}_{n}\text{, }\lambda\geq0,\label{Dual1}\\
&  \gamma\left(  v\right)  \geq l\left(  u,\beta\right)  -\lambda c\left(
u,v\right)  \text{ }\forall\text{ }\left(  u,v\right)  \in\mathcal{X}%
_{N}\times\mathcal{D}_{n}.\big\}\nonumber
\end{align}
%\begin{align*}
%&  \left.  \min\sum_{v\in\mathcal{D}_{N}}\gamma\left(  v\right)
%/n+\lambda\delta\right. \label{Dual1}\\
%\text{s.t.}&  \left.  \gamma\left(  v\right)  \geq l\left(  u,\beta\right)  -\lambda
%c\left(  u,v\right)  \text{ }\forall\text{ }\left(  u,v\right)  \in
%\mathcal{X}_{N}\times\mathcal{D}_{n},\right. \nonumber\\
%&  \left.  \gamma\left(  v\right)  \in\mathbb{R}\text{ }\forall\text{ }%
%v\in\mathcal{D}_{n}\text{, }\lambda\geq0\right.  .\nonumber
%\end{align*}
Maximizing over $u\in\mathcal{X}_{N}$ in the inequality constraint, for each
$v$, and using the fact that we are minimizing the objective function, we
obtain that (\ref{Dual1}) can be simplified to%
\[
\left.  E_{P_{n}}[\max_{u\in\mathcal{X}_{N}}\left\{  l\left(  u,\beta\right)
-\lambda c\left(  u,\left(  X,Y\right)  \right)  +\lambda\delta^{\ast
}\right\}  ]\right.  .
\]
Consequently, defining $\phi\left(  X,Y,\beta,\lambda\right)  = \max
_{u\in\mathcal{X}_{N}}\left\{  l\left(  u,\beta\right)  -\lambda c\left(
u,\left(  X,Y\right)  \right)  +\lambda\delta^{\ast}\right\}  $, we have that
(\ref{DRO_1}) is equivalent to
\begin{equation}
\min_{\lambda\geq0,\beta}\left.  E_{P_{n}}[\phi\left(  X,Y,\beta
,\lambda\right)  ]\right.  . \label{DRO_2}%
\end{equation}
Moreover, if we assume that $l\left(  u,\cdot\right)  $ is a convex function,
then we have that the mapping $\left(  \beta,\lambda\right)  \hookrightarrow
l\left(  u,\beta\right)  -\lambda c\left(  u,\left(  X,Y\right)  \right)
+\lambda\delta^{\ast}$is convex for each $u$ and therefore, $\left(
\beta,\lambda\right)  \hookrightarrow\phi\left(  X,Y,\beta,\lambda\right)  $,
being the maximum of convex mappings is also convex.

A natural approach consists in directly applying stochastic sub-gradient
descent (see \cite{boyd2004convex} and \cite{ram2010distributed}).
Unfortunately, this would involve performing the maximization over all
$u\in\mathcal{X}_{N}$ in each iteration. This approach could be prohibitively
expensive in typical machine learning applications where $N$ is large.

So, instead, we perform a standard smoothing technique, namely, we introduce
$\epsilon>0$ and define
\begin{align*}
\phi_{\epsilon}\left(  X,Y,\beta,\lambda\right)  = \lambda\delta^{\ast}
+\epsilon\log\big( \sum_{u\in\mathcal{X}_{N}}\exp\left(  \left\{  l\left(
u,\beta\right)  -\lambda c\left(  u,\left(  X,Y\right)  \right)  \right\}
/\epsilon\right)  \big) .
\end{align*}
It is easy to verify (using H\"{o}lder inequality) that $\phi_{\epsilon
}\left(  X,Y,\cdot\right)  $ is convex and it also follows that
\[
\phi\left(  X,Y,\beta,\lambda\right)  \leq\phi_{\epsilon}\left(
X,Y,\beta,\lambda\right)  \leq\phi\left(  X,Y,\beta,\lambda\right)
+\log(\left\vert \mathcal{X}_{N}\right\vert ) \epsilon.
\]
Hence, we can choose $\epsilon=O\left(  1/ \log N\right)  $ in order to
control the bias incurred by replacing $\phi$ by $\phi_{\epsilon}$. Then,
defining
\[
\tau_{\epsilon} \left(  X,Y,\beta,\lambda,u\right)  =\exp\left(  \left\{
l\left(  u,\beta\right)  -\lambda c\left(  u,\left(  X,Y\right)  \right)
\right\}  /\epsilon\right)  ,
\]
we have (assuming differentiability of $l\left(  u,\beta\right)  $) that%
\begin{align}
&  \nabla_{\beta}\phi_{\epsilon}\left(  X,Y,\beta,\lambda\right)  =\quad
\quad\quad\frac{\sum_{u\in\mathcal{X}_{N}}\tau_{\epsilon} \left(
X,Y,\beta,\lambda,u\right)  \nabla_{\beta}l\left(  u,\beta\right)  }%
{\sum_{v\in\mathcal{X}_{N}}\tau_{\epsilon} \left(  X,Y,\beta,\lambda,v\right)
},\label{Gradients}\\
&  \frac{\partial\phi_{\epsilon}\left(  X,Y,\beta,\lambda\right)  }%
{\partial\lambda} =\quad\quad\delta^{\ast}-\frac{\sum_{u\in\mathcal{X}_{N}
}\tau_{\epsilon} \left(  X,Y,\beta,\lambda,u\right)  c\left(  u,\left(
X,Y\right)  \right)  }{\sum_{v\in\mathcal{X}_{N}}\tau_{\epsilon} \left(
X,Y,\beta,\lambda,v\right)  }.\nonumber
\end{align}
In order to make use of the gradient representations (\ref{Gradients}) for the
construction of a stochastic gradient descent algorithm, we must construct
unbiased estimators for $\nabla_{\beta}\phi_{\epsilon}\left(  X,Y,\beta
,\lambda\right)  $ and $\partial\phi_{\epsilon}\left(  X,Y,\beta
,\lambda\right)  /\partial\lambda$, given $\left(  X,Y\right)  $. This can be
easily done if we assume that one can simulate directly $u\in\mathcal{X}_{N}$
with probability proportional to $\tau\left(  X,Y,\beta,\lambda,u\right)  $.
Because of the potential size of $\mathcal{X}_{N}$ and especially because such
distribution depends on $\left(  X,Y\right)  $ sampling with probability
proportional to $\tau_{\epsilon}\left(  X,Y,\beta,\lambda,u\right)  $ can be
very time-consuming.

So, instead, we apply a strategy discussed in \cite{blanchet2015unbiased}\ and
explained in Section 2.2.1. The proposed method produces random variables
$\Lambda\left(  X,Y,\beta,\lambda\right)  $ and $\Gamma\left(  X,Y,\beta
,\lambda\right)  $, which can be simulated easily by drawing i.i.d. samples
from the uniform distribution over $\mathcal{X}_{N}$, and such that
\begin{align*}
E\left(  \Lambda\left(  X,Y,\beta,\lambda\right)  |X,Y\right)  ={\partial
_{\lambda}}\phi_{\epsilon}\left(  X,Y,\beta,\lambda\right)  , \\E\left(
\Gamma\left(  X,Y,\beta,\lambda\right)  |X,Y\right)  =\nabla_{\beta}%
\phi_{\epsilon}\left(  X,Y,\beta,\lambda\right)  .
\end{align*}
Using this pair of random variables, then we apply the stochastic gradient
descent recursion%
\begin{align}
\label{SGD} &  \beta_{k+1} =\beta_{k}-\alpha_{k+1}\Gamma\left(  X_{k+1}%
,Y_{k+1},\beta_{k},\lambda_{k}\right)  ,\nonumber \\
&\lambda_{k+1} =\left(  \lambda
_{k}-\alpha_{k+1}\Lambda\left(  X_{k+1},Y_{k+1},\beta_{k},\lambda_{k}\right)
\right)  ^{+},
\end{align}
where learning sequence, $\alpha_{k} >0$ satisfies the standard conditions,
namely, $\sum_{k=1}^{\infty} \alpha_{k} = \infty${ and } $\sum_{k=1}^{\infty
}\alpha_{k}^{2} < \infty$, see \cite{shapiro2014lectures}.

We apply a technique from \cite{blanchet2015unbiased} to construct the random
variables $\Lambda$ and $\Gamma$, which originates from Multilevel Monte Carlo
introduced in \cite{giles2008multilevel}, and associated randomization methods
\cite{mcleish2011general},\cite{rhee2015unbiased}.

First, define $\bar{P}_{N}$ to be the uniform measure on $\mathcal{X}_{N}$ and
let $W$ be a random variable with distribution $\bar{P}_{N}$. Note that, given
$\left(  X,Y\right)  $,
\begin{align*}
&  \nabla_{\beta}\phi_{\epsilon}\left(  X,Y,\beta,\lambda\right)  =
\frac{E_{\bar{P}_{N}}\left(  \tau_{\epsilon} \left(  X,Y,\beta,\lambda
,W\right)  \nabla_{\beta}l\left(  W,\beta\right)  |\text{ }X,Y\right)
}{E_{\bar{P}_{N}}\left(  \tau_{\epsilon} \left(  X,Y,\beta,\lambda,W\right)
|\text{ }X,Y\right)  },\\
&  \partial_{\lambda}\phi_{\epsilon}\left(  X,Y,\beta,\lambda\right)
=\quad\delta^{\ast} - \frac{E_{\bar{P}_{N}}\left(  \tau_{\epsilon} \left(
X,Y,\beta,\lambda,W\right)  c\left(  W,\left(  X,Y\right)  \right)  |\text{
}X,Y\right)  }{E_{\bar{P}_{N}}\left(  \tau_{\epsilon} \left(  X,Y,\beta
,\lambda,W\right)  |\text{ }X,Y\right)  }.
\end{align*}
Note that both gradients can be written in terms of the ratios of two
expectations. The following results from \cite{blanchet2015unbiased} can be
used to construct unbiased estimators of functions of expectations. The
function of interest in our case is the ratio of expectations.

Let us define:
$h_{0}\left(  W\right)     =\tau_{\epsilon}\left(  X,Y,\beta,\lambda,W\right)$, 
$h_{1}\left(  W\right)     =h_{0}\left(  W\right)  c\left(  W,\left(
X,Y\right)  \right)  $ , and $h_{2}\left(  W\right)     =h_{0}\left(  W\right)  \nabla_{\beta}l\left(
W,\beta\right) $,
%\begin{align*}
%h_{0}\left(  W\right)   &  =\tau_{\epsilon}\left(  X,Y,\beta,\lambda,W\right)
%,\\
%h_{1}\left(  W\right)   &  =h_{0}\left(  W\right)  c\left(  W,\left(
%X,Y\right)  \right)  ,\\
%h_{2}\left(  W\right)   &  =h_{0}\left(  W\right)  \nabla_{\beta}l\left(
%W,\beta\right)  .
%\end{align*}
Then, we can write the gradient estimator as
\begin{align*}
\partial_{\lambda}\phi_{\epsilon}\left(  X,Y,\beta,\lambda\right)
=\frac{E_{\bar{P}_{N}}\left(  h_{1}\left(  W\right)  \text{ }|\text{
}X,Y\right)  }{E_{\bar{P}_{N}}\left(  h_{0}\left(  W\right)  \text{ }|\text{
}X,Y\right)  },\\ \text{ and }\nabla_{\beta}\phi_{\epsilon}\left(  X,Y,\beta
,\lambda\right)  =\frac{E_{\bar{P}_{N}}\left(  h_{2}\left(  W\right)  \text{
}|\text{ }X,Y\right)  }{E_{\bar{P}_{N}}\left(  h_{0}\left(  W\right)  \text{
}|\text{ }X,Y\right)  }.
\end{align*}

%\begin{align*}
%\partial_{\lambda}\phi_{\epsilon}\left(  X,Y,\beta,\lambda\right)  =\quad\quad
%g\left(  E_{\bar{P}_{N}}\left(  h_{0}\left(  W\right)  \text{ }|\text{
%}X,Y\right)  ,E_{\bar{P}_{N}}\left(  h_{1}\left(  W\right)  \text{ }|\text{
%}X,Y\right)  \right)  .
%\end{align*}
%Similarly,
%\begin{align*}
%\nabla_{\beta}\phi_{\epsilon}\left(  X,Y,\beta,\lambda\right)  = \quad
%g\left(  E_{\bar{P}_{N}}\left(  h_{0}\left(  W\right)  \text{ }|\text{
%}X,Y\right)  ,E_{\bar{P}_{N}}\left(  h_{2}\left(  W\right)  \text{ }|\text{
%}X,Y\right)  \right)  .
%\end{align*}
The procedure developed in \cite{blanchet2015unbiased} proceeds as follows.
First, define for a given $h\left(  W\right)  $, and $n\geq0$, the average
over odd and even labels to be
\[
\bar{S}_{2^{n}}^{E}\left(  h\right)  =\frac{1}{2^{n}}\sum_{i=1}^{2^{n}%
}h\left(  W_{2i}\right)  ,\text{ \ }\bar{S}_{2^{n}}^{O}\left(  h\right)
=\frac{1}{2^{n}}\sum_{i=1}^{2^{n}}h\left(  W_{2i-1}\right)  ,
\]
and the total average to be $\bar{S}_{2^{n+1}}\left(  h\right)  =\frac{1}%
{2}\left(  \bar{S}_{2^{n}}^{E}\left(  h\right)  +\bar{S}_{2^{n}}^{O}\left(
h\right)  \right)  $. We then state the following algorithm for sampling
unbiased estimators of $\partial_{\lambda}\phi_{\epsilon}\left(
X,Y,\beta,\lambda\right)  $ and $\nabla_{\beta}\phi_{\epsilon}\left(
X,Y,\beta,\lambda\right)  $ in Algorithm \ref{Algo_unbiased_gradient}.
\vspace{-5mm}
\begin{algorithm}[htb]

    \footnotesize
	\caption{Unbiased Gradient}
	\label{Algo_unbiased_gradient}
	\begin{algorithmic}[1]
		\State Given $\left( X,Y,\beta \right) $ the function outputs $\left( \Lambda
		,\Gamma \right) $ such that $E\left( \Lambda \right) =\partial _{\lambda
		}\phi _{\epsilon }\left( X,Y,\beta ,\lambda \right) $ and $E\left( \Gamma
		\right) =\nabla _{\beta }\phi _{\epsilon }\left( X,Y,\beta ,\lambda \right) $.
		\State {\bfseries Step1:}  Sample $G$ from geometric distribution with success parameter $%
		p_{G}=1-2^{-3/2}$.
		\State {\bfseries Step2:} Sample $W_{0},W_{1},...,W_{2^{G+1}}$ i.i.d. copies of $W$
		independent of $G$.
		\State {\bfseries Step3:} Compute
		\begin{align*}
			\Delta ^{\lambda } &= \frac{\bar{S}_{2^{G+1}}\left( h_{1}\right)}{\bar{S}%
						_{2^{G+1}}\left( h_{0}\right)} -
						\frac{1}{2}\left(
						\frac{\bar{S}_{2^{G+1}}^{O}\left( h_{1}\right)}{\bar{%
											S}_{2^{G+1}}^{O}\left( h_{0}\right)}
						+\frac{ \bar{S}_{2^{G}}^{E}\left( h_{1}\right)}{\bar{S}_{2^{G}}^{E}\left( h_{0}\right)}					
						\right),\\	
			\Delta ^{\beta } &= \frac{\bar{S}_{2^{G+1}}\left( h_{2}\right)}{\bar{S}%
									_{2^{G+1}}\left( h_{0}\right)} -
									\frac{1}{2}\left(
									\frac{\bar{S}_{2^{G+1}}^{O}\left( h_{2}\right)}{\bar{%
														S}_{2^{G+1}}^{O}\left( h_{0}\right)}
									+\frac{ \bar{S}_{2^{G}}^{E}\left( h_{2}\right)}{\bar{S}_{2^{G}}^{E}\left( h_{0}\right)}					
									\right)		.	
		\end{align*}
%		
%		\begin{eqnarray*}
%			\Delta ^{\lambda } &=& g\left( \bar{S}_{2^{G+1}}\left( h_{0}\right) ,\bar{S}%
%			_{2^{G+1}}\left( h_{1}\right) \right)  -\frac{ g\left( \bar{S}_{2^{G+1}}^{O}\left( h_{0}\right) ,\bar{%
%					S}_{2^{G+1}}^{O}\left( h_{1}\right) \right) +g\left( \bar{S}%
%				_{2^{G}}^{E}\left( h_{0}\right) ,\bar{S}_{2^{G}}^{E}\left( h_{1}\right)
%				\right)}{2}, \\
%			\Delta ^{\beta } &=&g\left( \bar{S}_{2^{G+1}}\left( h_{0}\right) ,\bar{S}%
%			_{2^{G+1}}\left( h_{2}\right) \right) -\frac{g\left( \bar{S}_{2^{G+1}}^{O}\left( h_{0}\right) ,\bar{%
%					S}_{2^{G+1}}^{O}\left( h_{1}\right) \right) +g\left( \bar{S}%
%				_{2^{G}}^{E}\left( h_{0}\right) ,\bar{S}_{2^{G}}^{E}\left( h_{1}\right)
%				\right)}{2} .
%		\end{eqnarray*}
		
		\State {\bfseries Output:}
		%		$\Lambda  = \delta^{\ast} -\frac{\Delta ^{\lambda }}{p_{G}\left( 1-p_{G}\right) ^{G}}%
		%					-g\left( h_{0}\left( W_{0}\right) ,h_{1}\left( W_{0}\right) \right)$ and $\quad\quad\quad\quad\quad\quad\quad\quad\quad\quad\quad$ $ \;\;$
		%					$\quad\quad\quad\quad\quad$ $\Gamma  =\frac{\Delta ^{\beta }}{p_{G}\left( 1-p_{G}\right) ^{G}}+g\left(
		%								h_{0}\left( W_{0}\right) ,h_{2}\left( W_{0}\right) \right)$.
		\begin{align*}
		\Lambda  = \delta^{\ast} -\frac{\Delta ^{\lambda }}{p_{G}\left( 1-p_{G}\right) ^{G}}%
					-\frac{ h_{1}\left( W_{0}\right)}{h_{0}\left( W_{0}\right)},\qquad					
		\Gamma  =\frac{\Delta ^{\beta }}{p_{G}\left( 1-p_{G}\right) ^{G}}+
			\frac{h_{2}\left( W_{0}\right)}{h_{0}\left( W_{0}\right)} 		.	
		\end{align*}
%		\begin{eqnarray*}
%			\Lambda  = \delta^{\ast} -\frac{\Delta ^{\lambda }}{p_{G}\left( 1-p_{G}\right) ^{G}}%
%			-g\left( h_{0}\left( W_{0}\right) ,h_{1}\left( W_{0}\right) \right) ,
%			\Gamma  =\frac{\Delta ^{\beta }}{p_{G}\left( 1-p_{G}\right) ^{G}}+g\left(
%			h_{0}\left( W_{0}\right) ,h_{2}\left( W_{0}\right) \right) .
%		\end{eqnarray*}
\vspace{-5mm}
	\end{algorithmic}
	
\end{algorithm}

\vspace{-5mm}
\section{Error Improvement of Our SSL-DRO Formulation
\label{Sec-Error-Improve}}
\vspace{-2mm}
Our goal in this section is to intuitively discuss why, owing to the inclusion
of the constraint $P\in\mathcal{P}\left(  \mathcal{X}_{N}\right)  $, we expect
desirable generalization properties of the SSL-DRO formulation (\ref{DRO_1}).
Moreover, our intuition suggests strongly why our SSL-DRO formulation should
possess better generalization performance than natural
supervised\ counterparts. We restrict the discussion for logistic regression
due to the simple form of regularization connection we will make in
\eqref{Eqn-Reg_LR-Rep}, however, the error improvement discussion should also
apply to general supervised learning setting.

As discussed in the\ Introduction using the game-theoretic interpretation of
(\ref{DRO_1}), by introducing $\mathcal{P}\left(  \mathcal{X}_{N}\right)  $,
the SSL-DRO formulation provides a mechanism for choosing $\beta$ which
focuses on potential out-of-sample scenarios which are more relevant based on
available evidence.

Suppose that the constraint $P\in\mathcal{P}\left(  \mathcal{X}_{N}\right)  $
was not present in the formulation. So, the inner maximization in
(\ref{DRO_1}) is performed over all probability measures $\mathcal{P}\left(
\mathbb{R}^{d+1}\right)  $ (supported on some subset of $\mathbb{R}^{d+1}$).
As indicated earlier, we assume that $l\left(  X,Y;\cdot\right)  $ is strictly
convex and differentiable, so the first order optimality condition
$E_{P}\left(  \nabla_{\beta}l\left(  X,Y;\beta\right)  \right)  =0$
characterizes the optimal choice of $\beta$ assuming the validity of the
probabilistic model $P$. It is natural to assume that there exists an actual
model underlying the generation of the training data, which we denote as
$P_{\infty}$. Moreover, we may also assume that there exists a unique
$\beta^{\ast}$ such that $E_{P_{\infty}}\left(  \nabla_{\beta}l\left(
X,Y;\beta^{\ast}\right)  \right)  =0$.

The set $\mathcal{M}\left(  \beta_{\ast}\right)  =\mathcal{\{}P\in\mathcal{P}\left(
\mathbb{R}^{d+1}\right)  :E_{P}\left(  \nabla_{\beta}l\left(  X,Y;\beta^{\ast
}\right)  \right)  =0\mathcal{\}}$
%\[
%\mathcal{M}\left(  \beta_{\ast}\right)  =\mathcal{\{}P\in\mathcal{P}\left(
%\mathbb{R}^{d+1}\right)  :E_{P}\left(  \nabla_{\beta}l\left(  X,Y;\beta^{\ast
%}\right)  \right)  =0\mathcal{\}}%
%\]
corresponds to the family of all probability models which correctly estimate
$\beta^{\ast}$. Clearly, $P_{\infty}\in\mathcal{M}\left(  \beta_{\ast}\right)
$, whereas, typically, $P_{n}\notin\mathcal{M}\left(  \beta_{\ast}\right)  $.
Moreover, if we write $\mathcal{X}_{\infty}=$ $supp\left(  P_{\infty}\right)
$ we have that
\[
P_{\infty}\in m\left(  N,\beta^{\ast}\right)  :=\mathcal{\{}P\in
\mathcal{P}\left(  \mathcal{X}_{\infty}\right)  :\quad E_{P}\left(
\nabla_{\beta}l\left(  X,Y;\beta^{\ast}\right)  \right)  =0\mathcal{\}\subset
M}\left(  \beta_{\ast}\right)  .
\]
Since $\mathcal{X}_{N}$ provides a sketch of $\mathcal{X}_{\infty}$, then we
expect to have that the extremal (i.e. worst case) measure, denoted by
$P_{N}^{\ast}$, will be in some sense a better description of $P_{\infty}$.
\begin{figure}[pth]
\vspace{-5mm}
\par
\begin{center}
\centerline{\includegraphics[width=0.6 \textwidth]{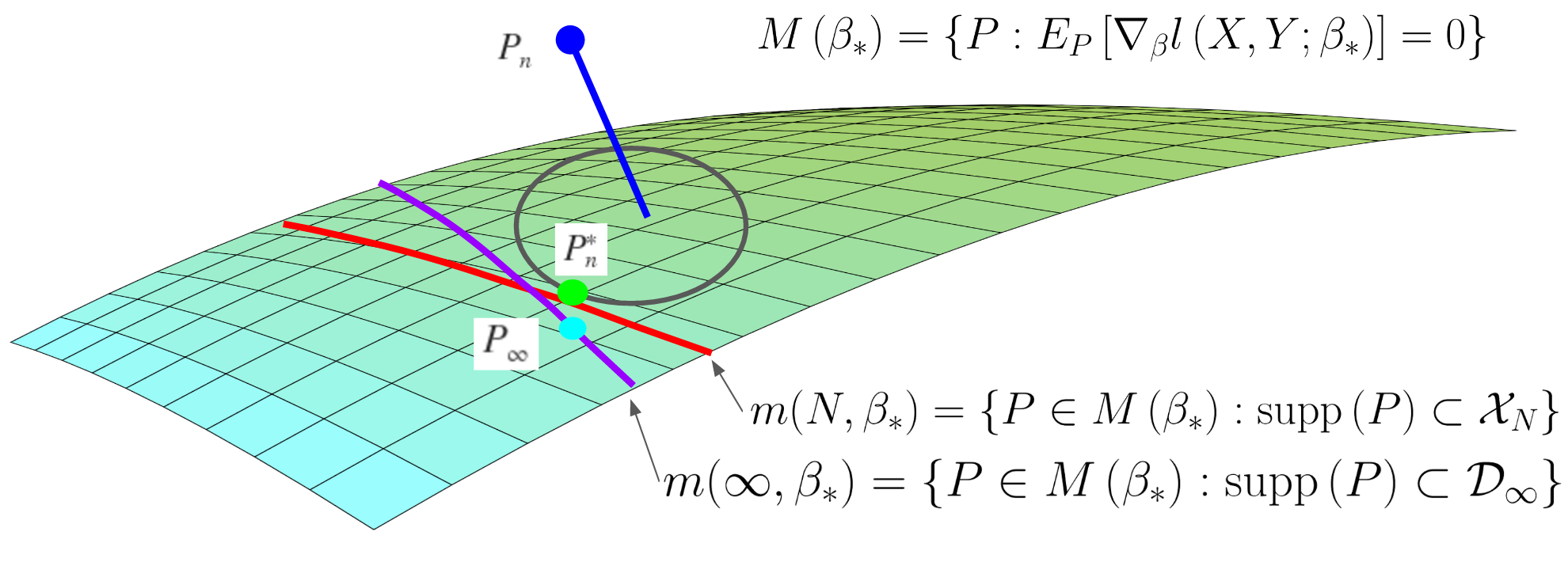}}
\end{center}
\par
\v -0.2in\caption{Pictorial representation of the role that the support
constraint plays in the SSL-DRO approach and how its presence enhances the
out-of-sample performance.}%
\label{Fig_DRO_intui3}%
\vspace{-5mm}
\end{figure}
%\begin{figure}[pth]
%\begin{center}
%\includegraphics[width=\textwidth]{General_plot}
%\end{center}
%\caption{(a): Idealization of the way in which the unlabeled
%		predictive variables provide a proxy for an underlying lower dimensional
%		manifold. Large red dots represent labeled instances and small blue dots
%		represent unlabeled instances. (b)Pictorial representation of the role that the support
%		constraint plays in the SSL-DRO approach and how its presence
%		enhances the out-of-sample performance.}%
%\label{Fig_DRO_intui0}%
%\end{figure}

Figure \ref{Fig_DRO_intui3} provides a\ pictorial representation of the
previous discussion. In the absence of the constraint $P\in\mathcal{P}\left(
\mathcal{X}_{N}\right)  $, the extremal measure chosen by nature can be
interpreted as a projection of $P_{n}$ onto $\mathcal{M}\left(  \beta_{\ast
}\right)  $. In the presence of the constraint $P\in\mathcal{P}\left(
\mathcal{X}_{N}\right)  $, we can see that $P_{N}^{\ast}$ may bring the
learning procedure closer to $P_{\infty}$. Of course, if $N$ is not large
enough, the schematic may not be valid because one may actually have $m\left(
N,\beta^{\ast}\right)  = \varnothing$.

The previous discussion is useful to argue that our SSL-DRO formulation should
be superior to the DRO\ formulation which is not informed by the unlabeled
data. But this comparison may not directly apply to alternative supervised
procedures that are mainstream in machine learning, which should be considered
as the natural benchmark to compare with. Fortunately, replacing the
constraint that $P\in\mathcal{P}\left(  \mathcal{X}_{N}\right)  $ by
$P\in\mathcal{P}\left(  \mathbb{R}^{d+1}\right)  $ in the DRO formulation
recovers exactly supervised learning algorithms such as regularized logistic regression.

Recall from \cite{blanchet2016robust} that if $l\left(  x,y,\beta\right)
=\log(1+\exp(-y\cdot\beta^{T}x))$ and if we define
\[
c\left(  (x,y),(x^{\prime},y^{\prime})\right)  = \Vert x-x^{\prime}\Vert_{q}
I(y=y^{\prime}) + \infty I(y\neq y^{\prime}),
\]
for $q\geq1$ then, according to Theorem 3 in \cite{blanchet2016robust}, we
have that
\begin{equation}
\label{Eqn-Reg_LR-Rep}\min_{\beta}\max_{D_{c}\left(  P,P_{n}\right)  \leq
\bar{\delta}}E_{P}[l\left(  X,Y,\beta\right)  ]=\min_{\beta\in\mathbb{R}^{d}%
}\left\{  E_{P_{n}}\left[  l\left(  X,Y,\beta\right)  \right]  +\bar{\delta
}\left\Vert \beta\right\Vert _{p}\right\}  ,
\end{equation}
where $q$ satisfies $1/p+1/q=1$. Formulation (\ref{DRO_1}) is, therefore, the
natural SSL extension of the standard regularized logistic regression estimator.

We conclude that, for logistic regression, SSL-DRO as formulated in
(\ref{DRO_1}), is a natural SSL extension of the standard regularized logistic
regression estimator, which would typically induce superior generalization
abilities over its supervised counterparts, and similar discussion should
apply to most supervised learning methods.
\vspace{-3mm}
\section{Numerical Experiments \label{Sec_Numerical}}
\vspace{-3mm}
We proceed to numerical experiments to verify the performance of our SSL-DRO
method empirically using six binary classification real data sets from UCI
machine learning data base \cite{Lichman:2013}.

We consider our SSL-DRO formulation based on logistic regression and compare
with other state-of-the-art logistic regression based SSL algorithms, entropy
regularized logistic regression with $L_{1}$ regulation (ERLRL1)
\cite{grandvalet2005semi} and regularized logistic regression based
self-training (STLRL1) \cite{li2008self}. In addition, we also compare with
its supervised counterpart, which is regularized logistic regression (LRL1).
For each iteration of a data set, we randomly split the data into labeled
training, unlabeled training and testing set, we train the models on training
sets and evaluate the testing error and accuracy with testing set. We report
the mean and standard deviation for training and testing error using
log-exponential loss and the average testing accuracy, which are calculated
via $200$ independent experiments for each data set. We summarize the detailed
results, the basic information of the data sets, and our data split setting in
Table \ref{Table-Real}.

We can observe that our SSL-DRO method has the potential to improve upon these
state-of-the-art SSL algorithms.

\begin{table}[th]
\footnotesize
\label{Table-Real}
\centering
\begin{tabular}
[c]{cccccccc}
&  & Breast Cancer & qsar & Magic & Minibone & Spambase & \\\hline
\multicolumn{1}{l}{{LRL1}} & Train & $.185\pm.123$ &
$.614\pm.038$ & $.548\pm.087$ & $.401\pm.167$ & $.470 \pm.040$ & \\
\multicolumn{1}{l}{} & Test & $.428\pm.338$ & $.755\pm.019$ & $.610\pm.050$ &
$.910\pm.131$ & $.588 \pm.141$ & \\
\multicolumn{1}{l}{} & Accur & $.929\pm.023$ & $.646\pm.036$ & $.665\pm.045$ &
$.717\pm.041$ & $.811\pm.034$ & \\\hline
\multicolumn{1}{l}{{ERLRL1}} & Train & $.019\pm.010$ &
$.249\pm.050$ & $2.37\pm.987$ & $.726\pm.353$ & $.008 \pm.028$ & \\
\multicolumn{1}{l}{} & Test & $.265\pm.146$ & $.720\pm.029$ & $4.28\pm1.51$ &
$1.98\pm.678$ & $.505 \pm.108$ & \\
\multicolumn{1}{l}{} & Accur & $.944\pm.018$ & $.731\pm.026$ & $.721\pm.056$ &
$.708\pm.071$ & $.883\pm.018$ & \\\hline
\multicolumn{1}{l}{{STLRL1}} & Train & $.089\pm.019$ &
$.498\pm.120$ & $3.05\pm.987$ & $1.50\pm.706$ & $.370\pm.082$ & \\
\multicolumn{1}{l}{} & Test & $.672\pm.034$ & $2.37\pm.860$ & $8.03\pm1.51$ &
$4.81\pm.732$ & $1.47\pm.316$ & \\
\multicolumn{1}{l}{} & Accur & $.955\pm.023$ & $.694\pm.038$ & $.692\pm.056$ &
$.704 \pm.033$ & $.843 \pm.023$ & \\\hline
\multicolumn{1}{l}{{DROSSL}} & Train & $.045\pm.023$ &
$.402\pm.039$ & $.420\pm.075$ & $.287\pm.047$ & $.221\pm.028$ & \\
\multicolumn{1}{l}{} & Test & $.120\pm.029$ & $.555\pm.025$ & $.561\pm.039$ &
$.609\pm.054$ & $.333 \pm.012$ & \\
\multicolumn{1}{l}{} & Accur & $.956\pm.016$ & $.734\pm.025$ & $.733\pm.034$ &
$.710\pm.032$ & $.892 \pm.009$ & \\\hline\hline
\multicolumn{2}{l}{Num Predictors} & $30$ & $30$ & $10$ & $20$ & $56$ & \\
\multicolumn{2}{l}{Labeled Size} & $40$ & $80$ & $30$ & $30$ & $150$ & \\
\multicolumn{2}{l}{Unlabeled Size} & $200$ & $500$ & $9000$ & $5000$ & $1500$
& \\
\multicolumn{2}{l}{Testing Size} & $329$ & $475$ & $9990$ & $125034$ & $2951$
&
\end{tabular}
\caption{Numerical experiments for real data sets.}%
\vspace{-5mm}
\end{table}
\vspace{-3mm}
\section{Discussion on the Size of the Uncertainty Set\label{Sec_del_star}}
\vspace{-3mm}
One of the advantages of DRO\ formulations such as \eqref{DRO_1} and
\eqref{Eqn-Reg_LR-Rep} is that they lead to a natural criterion for the
optimal choice of the parameter $\delta^{\ast}$ or, in the case of
\eqref{Eqn-Reg_LR-Rep}, the choice of $\bar{\delta}$ (which incidentally
corresponds to the regularization parameter). The optimality criterion that we
use to select the size of $\delta^{\ast}$ is motivated by Figure
\ref{Fig_DRO_intui3}.

First, interpret the uncertainty set
\[
\mathcal{U}_{\delta}\left(  P_{n},\mathcal{X}_{N}\right)  =\{P\in
\mathcal{P}\left(  \mathcal{X}_{N}\right)  :D_{c}\left(  P,P_{n}\right)
\leq\delta\}
\]
as the set of plausible models which are consistent with the empirical
evidence encoded in $P_{n}$ and $\mathcal{X}_{N}$. Then, for every plausible
model $P$, we can compute $\beta\left(  P\right)  =\arg\min_{\beta} E_{P}[l\left(  X,Y,\beta\right)  ]$, 
%\[
%\beta\left(  P\right)  =\arg\min_{\beta} E_{P}[l\left(  X,Y,\beta\right)  ]
%\]
and therefore the set 
$\Lambda_{\delta}\left(  P_{n},\mathcal{X}_{N}\right)  =\{\beta\left(
P\right)  =\arg\min E_{P}[l\left(  X,Y,\beta\right)  ]:P\in\mathcal{U}%
_{\delta}\left(  P_{n},\mathcal{X}_{N}\right)  \}$
%\[
%\Lambda_{\delta}\left(  P_{n},\mathcal{X}_{N}\right)  =\{\beta\left(
%P\right)  =\arg\min E_{P}[l\left(  X,Y,\beta\right)  ]:P\in\mathcal{U}%
%_{\delta}\left(  P_{n},\mathcal{X}_{N}\right)  \}
%\]
can be interpreted as a confidence region. It is then natural to select a
confidence level $\alpha\in\left(  0,1\right)  $ and compute $\delta^{\ast
}:=\delta_{N,n}^{\ast}$ by solving%
\begin{equation}
\label{Eqn-SSL-RWP-mot}\min\{\delta:P\left(  \beta^{\ast}\in\Lambda_{\delta
}\left(  P_{n},\mathcal{X}_{N}\right)  \right)  \geq1-\alpha\}.
\end{equation}
Similarly, for the supervised version, we can select $\bar{\delta}=\bar
{\delta}_{n}$ by solving
\begin{equation}
\label{Eqn-Supervised-RWP-mot}\min\{\delta:P\left(  \beta^{\ast}\in
\Lambda_{\delta}\left(  P_{n},\mathbb{R}^{d+1}\right)  \right)  \geq
1-\alpha\}.
\end{equation}

It is easy to see that $\bar{\delta}_{n}\leq\delta_{N,n}^{\ast}$. Now, we let
$N=\gamma n$ for some $\gamma>0$ and consider $\delta_{N,n}^{\ast}$,
$\bar{\delta}_{n}$ as $n\rightarrow\infty$. This analysis is relevant because
we are attempting to sketch $supp\left(  P_{\infty}\right)  $ using the set
$\mathcal{X}_{N}$, while considering large enough plausible variations to be
able to cover $\beta^{\ast}$ with $1-\alpha$ confidence.
\newline
More precisely, following the discussion in \cite{blanchet2016robust} for the
supervised case in finding $\bar{\delta}_{n}$ in \eqref{Eqn-SSL-RWP-mot} using
Robust Wasserstein Profile (RWP) function, solving
\eqref{Eqn-Supervised-RWP-mot} for $\delta_{N,n}^{\ast}$ is equivalent to
finding the $1-\alpha$ quantile of the asymptotic distribution of the RWP
function, defined as
\begin{align}
R_{n}\left(  \beta\right)  =\min_{\pi} &  \big\{\sum_{u\in\mathcal{X}_{n}}%
\sum_{v\in\mathcal{D}_{n}}c(u,v)\pi(u,v),\sum_{u\in\mathcal{X}_{n}}%
\pi(u,v)=\frac{1}{n},\forall v\mathcal{D}_{n},\label{Eqn-RWP}\\
&  \qquad\pi\subset\mathcal{P}\left(  \mathcal{X}_{n}%
\times\mathcal{D}_{n}\right)  ,\sum_{u\in\mathcal{X}_{n}}\sum_{v\in
\mathcal{D}_{n}}\nabla_{\beta}l\left(  u;\beta\right)  \pi
(u,v)=0.\big\}.\nonumber
\end{align}
The RWP function is the distance, measured by the optimal transport cost
function, between the empirical distribution and the manifold of probability
measures for which $\beta_{\ast}$ is the optimal parameter. A pictorial
representation is given in Figure \ref{Fig_DRO_intui3}. Additional discussion on the
RWP function and its interpretations can be found in
\cite{blanchet2016robust,blanchet2016sample}.

In the setting of the DRO formulation for \eqref{Eqn-Reg_LR-Rep} it is shown
in \cite{blanchet2016robust}, that $\bar{\delta}_{n}=O\left(  n^{-1}\right)  $
for \eqref{Eqn-Reg_LR-Rep} as $n\rightarrow\infty$. Intuitively, we expect
that if the predictive variables possess a positive density supported in a
lower dimensional manifold of dimension $\bar{d}<d$, then sketching
$supp\left(  P_{\infty}\right)  $ with $O\left(  n\right)  $ data points will
leave relatively large portions of the manifold unsampled (since, on average,
$O\left(  n^{\bar{d}}\right)  $ sampled points are needed to be within
distance $O\left(  1/n\right)  $ of a given point in box of unit size in
$\bar{d}$ dimensions). The optimality criterion will recognize this type of
discrepancy between $\mathcal{X}_{N}$ and $supp\left(  P_{\infty}\right)  $.
Therefore, we expect that $\delta_{\gamma n,n}^{\ast}$ will converge to zero
at a rate which might deteriorate slightly as $\bar{d}$ increases. 

This intuition is given rigorous support in Theorem \ref{SoS_theorem_LM} for
linear regression with square loss function and $L_{2}$ cost
function for DRO. In turn, Theorem \ref{SoS_theorem_LM} follows as a corollary
to the results in \cite{blanchet2016sample}. To make our paper self-contained, we have the detailed assumptions and a sketch of proof in the appendix.
\vspace{-3mm}
\begin{theorem}{1}\label{SoS_theorem_LM}
Assume the linear regression model $Y = \beta^{\ast}X + e$ with square loss function, i.e.
$l\left(X,X;\beta\right) = \left(Y-\beta^{T}X\right)^{2}$, and transport cost 
	\[
		c\left( \left( x,y\right) ,\left( x^{\prime },y^{\prime }\right) \right)
		= 		\left\Vert x-x^{\prime }\right\Vert _{2}^{2}I_{y=y^{\prime}} + \infty I_{y\neq y^{\prime}}.
	\]
	Assume $N=\gamma n$ and under mild assumptions on $\left(X,Y\right)$, if we denote
	$\tilde{Z}\sim\mathcal{N}\left(0,E[V_{1}]\right)$, we have:
		\begin{itemize}
			\item When $d=1$, $nR_{n}(\beta_{\ast} )\Rightarrow \kappa _{1}\chi _{1}^{2}.$%
		%	\begin{equation*}
		%	nR_{n}(\beta_{\ast} )\Rightarrow \kappa _{1}\chi _{1}^{2}.
		%	\end{equation*}
			
			\item When $d=2$, $nR_{n}(\beta_{\ast} )\Rightarrow F_{2}\left(\tilde{Z}\right),$
			%\begin{equation*}
			%nR_{n}(\beta_{\ast} )\Rightarrow F_{2}\left(\tilde{Z}\right),
			%\end{equation*}
			where $F_{2}(\cdot)$ is a continuous function and $F_{2}\left(z\right)
			= O(\Vert z \Vert_{2}^{2})$ as $\Vert z \Vert_{2} \to \infty $.
			
			\item When $d\geq 3$, $n^{1/2+\frac{3}{2d+2}}R_{n}(\beta_{\ast} )\Rightarrow
			F_{d}\left(\tilde{Z}\right),$
			%\begin{equation*}
			%n^{1/2+\frac{3}{2d+2}}R_{n}(\beta_{\ast} )\Rightarrow
			%F_{d}\left(\tilde{Z}\right),
			%\end{equation*}
			where $F_{d}\left(\cdot\right)$ is a continuous function (depending on $d$) and $F_{d}\left(z\right) = O\big(
				\left\Vert {z}\right\Vert_{2}^{d/2+1}
			\big)$.
		\end{itemize}
\end{theorem}

It is shown in Theorem \ref{SoS_theorem_LM} for SSL linear regression that
when $q=2$, $\delta_{\gamma n,n}^{\ast}=O\big(n^{-1/2-3/(2\bar{d}+2)}\big)$
for $\bar{d}\geq3$, and $\delta_{\gamma n,n}^{\ast}=O\left(  n^{-1}\right)  $
for $\bar{d}=1,2$. A similar argument can be made for logistic regression as
well. We believe that this type of analysis and its interpretation is of
significant interest and we expect to report a more complete picture in the
future, including the case $q\geq1$ (which we believe should obey the same scaling).
\vspace{-3mm}
\section{Conclusions\label{Sec-Con}}
\vspace{-3mm}
We have shown that our SSL-DRO, as a semi-supervised method, is able to enhance the generalization predicting power versus its supervised counterpart. Our
numerical experiments show superior performance of our SSL-DRO method when
compared to state-of-the-art SSL algorithms such as ERLRL1 and STLRL1. We
would like to emphasize that our SSL-DRO method is not restricted to linear
and logistic regressions. As we can observe from the DRO formulation and the
algorithm. If a learning algorithm has an accessible loss function and the
loss gradient can be computed, we are able to formulate the SSL-DRO problem
and benefit from unlabeled information. Finally, we discussed a stochastic
gradient descent technique for solving DRO problems such as (\ref{DRO_1}),
which we believe can be applied to other settings in which the gradient is a
non-linear function of easy-to-sample expectations.
%In the statement of the previous result, the symbol $W_{n}\lesssim_{D}W$
%means
%\[
%\overline{\lim}_{n\rightarrow\infty}P\left(  W_{n}>x\right)  \leq P\left(
%W>x\right)
%\]
%for any $x$. The previous result

%\section{Section Title}
%Main contents here.
%\subsection{Subsection Title}
%A figure in Fig.~\ref{fig:spiral}. Please use high quality graphics for your camera-ready submission -- if you can use a vector graphics format such as \texttt{.eps} or \texttt{.pdf}.
%\begin{figure}[htp]
%\begin{center}
%\includegraphics[width=0.5\textwidth]{spiral.eps}
%\caption{A spiral.}\label{fig:spiral}
%\end{center}
%\end{figure}
%An example of citation~\cite{DBLP:conf/acml/2009}.

%\acks{Acknowledgements should go at the end, before appendices and references.}

%\bibliographystyle{plain}
\bibliographystyle{plain}
\bibliography{SOS_semisupervised}

\begin{thebibliography}{10}

\bibitem{balsubramani2015scalable}
Akshay Balsubramani and Yoav Freund.
\newblock Scalable semi-supervised aggregation of classifiers.
\newblock In {\em NIPS}, pages 1351--1359, 2015.

\bibitem{bertsimas2011theory}
Dimitris Bertsimas, David Brown, and Constantine Caramanis.
\newblock Theory and applications of robust optimization.
\newblock {\em SIAM review}, 53(3):464--501, 2011.

\bibitem{bertsimas2013data}
Dimitris Bertsimas, Vishal Gupta, and Nathan Kallus.
\newblock Data-driven robust optimization.
\newblock {\em arXiv preprint arXiv:1401.0212}, 2013.

\bibitem{blanchet2015unbiased}
Jose Blanchet and Peter Glynn.
\newblock Unbiased {M}onte {C}arlo for optimization and functions of
  expectations via multi-level randomization.
\newblock In {\em Proceedings of the 2015 Winter Simulation Conference}, pages
  3656--3667. IEEE Press, 2015.

\bibitem{blanchet2016sample}
Jose Blanchet and Yang Kang.
\newblock Sample out-of-sample inference based on wasserstein distance.
\newblock {\em arXiv preprint arXiv:1605.01340}, 2016.

\bibitem{blanchet2016robust}
Jose Blanchet, Yang Kang, and Karthyek Murthy.
\newblock Robust wasserstein profile inference and applications to machine
  learning.
\newblock {\em arXiv preprint}, 2016.

\bibitem{blum2001learning}
Avrim Blum and Shuchi Chawla.
\newblock Learning from labeled and unlabeled data using graph mincuts.
\newblock 2001.

\bibitem{boyd2004convex}
Stephen Boyd and Lieven Vandenberghe.
\newblock {\em Convex optimization}.
\newblock Cambridge university press, 2004.

\bibitem{chapelle2009semi}
Olivier Chapelle, Bernhard Scholkopf, and Alexander Zien.
\newblock Semi-supervised learning.
\newblock {\em IEEE Transactions on Neural Networks}, 20(3):542--542, 2009.

\bibitem{giles2008multilevel}
Michael Giles.
\newblock Multilevel {M}onte {C}arlo path simulation.
\newblock {\em Operations Research}, 56(3), 2008.

\bibitem{grandvalet2005semi}
Yves Grandvalet and Yoshua Bengio.
\newblock Semi-supervised learning by entropy minimization.
\newblock In {\em Advances in {NIPS}}, pages 529--536, 2005.

\bibitem{li2008self}
Yuanqing Li, Cuntai Guan, Huiqi Li, and Zhengyang Chin.
\newblock A self-training semi-supervised svm algorithm and its application in
  an eeg-based brain computer interface speller system.
\newblock {\em Pattern Recognition Letters}, 29(9):1285--1294, 2008.

\bibitem{Lichman:2013}
Moshe. Lichman.
\newblock {UCI} machine learning repository, 2013.

\bibitem{loog2016contrastive}
Marco Loog.
\newblock Contrastive pessimistic likelihood estimation for semi-supervised
  classification.
\newblock {\em IEEE transactions on pattern analysis and machine intelligence},
  38(3):462--475, 2016.

\bibitem{luenberger1973introduction}
David~G Luenberger.
\newblock {\em Introduction to linear and nonlinear programming}, volume~28.
\newblock Addison-Wesley Reading, MA, 1973.

\bibitem{mcleish2011general}
Don McLeish.
\newblock A general method for debiasing a {M}onte {C}arlo estimator.
\newblock {\em Monte Carlo Meth. and Appl.}, 17(4):301--315, 2011.

\bibitem{ram2010distributed}
Sundhar Ram, Angelia Nedi{\'c}, and Venugopal Veeravalli.
\newblock Distributed stochastic subgradient projection algorithms for convex
  optimization.
\newblock {\em Journal of optimization theory and applications},
  147(3):516--545, 2010.

\bibitem{rhee2015unbiased}
Chang-han Rhee and Peter Glynn.
\newblock Unbiased estimation with square root convergence for {SDE} models.
\newblock {\em Operations Research}, 63(5):1026--1043, 2015.

\bibitem{rubner2000earth}
Yossi Rubner, Carlo Tomasi, and Leonidas Guibas.
\newblock The earth mover's distance as a metric for image retrieval.
\newblock {\em International journal of computer vision}, 2000.

\bibitem{shafieezadeh2015distributionally}
Soroosh Shafieezadeh-Abadeh, Peyman~Mohajerin Esfahani, and Daniel Kuhn.
\newblock Distributionally robust logistic regression.
\newblock In {\em NIPS}, pages 1576--1584, 2015.

\bibitem{shapiro2014lectures}
Alexander Shapiro, Darinka Dentcheva, et~al.
\newblock {\em Lectures on stochastic programming: modeling and theory},
  volume~16.
\newblock Siam, 2014.

\bibitem{villani2008optimal}
C{\'e}dric Villani.
\newblock {\em Optimal transport: old and new}, volume 338.
\newblock Springer Science \& Business Media, 2008.

\bibitem{xu2009robust}
Huan Xu, Constantine Caramanis, and Shie Mannor.
\newblock Robust regression and lasso.
\newblock In {\em Advances in Neural Information Processing Systems}, pages
  1801--1808, 2009.

\bibitem{zhu2005semi}
Xiaojin Zhu, John Lafferty, and Ronald Rosenfeld.
\newblock {\em Semi-supervised learning with graphs}.
\newblock Carnegie Mellon University, 2005.

\end{thebibliography}
\newpage
\appendix

\section{Supplementary Material: Technical Details for Theorem \ref{SoS_theorem_LM}}

\label{Sec-Appendix-Tech} In this supplementary appendix, we first state the general
assumptions to guarantee the validity of the asymptotically optimal selection
for the distributional uncertainty size in Section
\ref{Sec-Assumptions-SoS-Theorem}. In Section \ref{Sec-Supp-Th} and \ref{Sec-Supp-Proof} we revisit 
Theorem \ref{SoS_theorem_LM} and provide a more detailed proof. 

\subsection{Assumptions of Theorem \ref{SoS_theorem_LM}}

\label{Sec-Assumptions-SoS-Theorem} For linear regression model, let us assume
we have a collection of labeled data $\mathcal{D}_{n}=\left\{  \left(
X_{i},Y_{i}\right)  \right\}  _{i=1}^{n}$ and a collection of unlabeled data
$\left\{  X_{i}\right\}  _{i=n+1}^{N}$. We consider the set $\mathcal{X}%
_{N}=\left\{  X_{i}\right\}  _{i=1}^{N}\times\left\{  Y_{i}\right\}
_{i=1}^{n}$, to be the cross product of all the predictors from labeled and
unlabeled data and the labeled responses. In order to have proper asymptotic
results holds for the RWP function, we require some mild assumptions on the
density and moments of $\left(  X,Y\right)  $ and estimating equation
$\nabla_{\beta}l\left(  X,Y;\beta\right)  =\left(  Y-\beta_{\ast}^{T}\right)
X$. We state them explicitly as follows:

\textbf{A)} We assume the predictors $X_{i}$'s for the labeled and unlabeled
data are i.i.d. from the same distribution with positive differentiable
density $f_{X}(\cdot)$ with bounded bounded gradients.

\textbf{B)} We assume the $\beta_{\ast}\in\mathbb{R}^{d}$ is the true
parameter and under null hypothesis of the linear regression model satisfying
$Y=\beta_{\ast}^{T}X+e$, where $e$ is a random error independent of $X$.

\textbf{C)} We assume $E\left[  X^{T}X\right]  $ exists and is positive
definite and $E\left[  e^{2}\right]  <\infty$.

\textbf{D)} For the true model of labeled data, we have $E_{P_{\ast}}\left[
X\left(  Y-\beta_{\ast}^{T}X\right)  \right]  =0$ (where $P_{\ast}$ denotes
the actual population distribution which is unknown).

The first two assumptions, namely Assumption A and B, are elementary
assumptions for linear regression model with an additive independent random
error. The requirements for the differentiable positive density for the
predictor $X$, is because when $d\geq3$, the density function appears in the
asymptotic distribution. Assumption C is a mild requirement on the moments
exist for predictors and error, and Assumption D is to guarantee true
parameter $\beta_{\ast}$ could be characterized via first order optimality
condition, i.e. the gradient of the square loss function. Due to the simple
structure of the linear model, with the above four assumptions, we can prove
Theorem \ref{SoS_theorem_LM} and we show a sketch in the following subsection.

\subsection{Revisit Theorem \ref{SoS_theorem_LM}}

\label{Sec-Supp-Th} In this section, we revisit the asymptotic result for
optimally choosing uncertainty size for semi-supervised learning for the
linear regression model. We assume that, under the null hypothesis,
$Y=\beta_{\ast}^{T}X+e$, where $X\in\mathbb{R}^{d}$ is the predictors, $e$ is
independent of $X$ as random error, and $\beta_{\ast}\in\mathbb{R}^{d}$ is the
true parameter. We consider the square loss function and assume that
$\beta_{\ast}$ is the minimizer to the square loss function, i.e.
\[
\beta_{\ast}=\arg\min_{\beta}E\left[  \left(  Y-\beta^{T}X\right)
^{2}\right]  .
\]
If we can assume the second-moment exists for $X$ and $e$, then we can switch
the order of expectation and derivative w.r.t. $\beta$, then optimal $\beta$
could be uniquely characterized via the first order optimality condition,
\[
E\left[  X\left(  Y-\beta_{\ast}^{T}X\right)  \right]  =0.
\]
As we discussed in Section \ref{Sec_del_star}, the optimal distributional
uncertainty size $\delta_{n,N}^{\ast}$ at confidence level $1-\alpha$, is
simply the $1-\alpha$ quantile of the RWP function defined in \eqref{Eqn-RWP}.
In turn, the asymptotic limit of the RWP function is characterized in Theorem
1, which we restate more explicitly here.

\textbf{Theorem 1}[Restate of Theorem \ref{SoS_theorem_LM} in Section
	\ref{Sec_del_star}]
	For linear regression model we defined above and square loss function, if we take
	cost function for DRO formulation to be
	\[
			c\left( \left( x,y\right) ,\left( x^{\prime },y^{\prime }\right) \right)
			= 		\left\Vert x-x^{\prime }\right\Vert _{2}^{2}I_{y=y^{\prime}} + \infty I_{y\neq y^{\prime}}.
		\]
	If we assume Assumptions A,B, and D stated in Section \ref{Sec-Assumptions-SoS-Theorem}
	to be true and number of unlabeled data satisfing $N=\gamma n$. Furthermore, let us
	denote: $V_{i} = \left(e_{i}I -X_{i}\beta_{\ast}^{T} \right)\left(e_{i}I -\beta_{\ast}X_{i}^{T} \right)$, where $e_{i} = Y_{i}-\beta_{\ast}^{T}X_{i}$ being
	the residual under the null hypothesis. Then, we have:
			\begin{itemize}
				\item When $d=1$,%
				\begin{equation*}
				nR_{n}(\beta_{\ast} )\Rightarrow \frac{E\left[X_{1}^{2}e_{1}^{2}\right]}{
								E\left[\left(e_{1}-\beta_{\ast}^{T}X_{1}\right)^{2}\right]}\chi _{1}^{2}.
				\end{equation*}
				
				\item When $d=2$,
				\begin{equation*}
				nR_{n}(\beta_{\ast} )\Rightarrow 2 \tilde{\zeta}(\tilde{Z})^{T}\tilde{Z} -
						\tilde{\zeta}\left(\tilde{Z}\right)^{T}  \tilde{G}_{2}\left(\tilde{\zeta}\left(\tilde{Z}\right) \right)\tilde{\zeta}\left(\tilde{Z}\right) ,
				\end{equation*}
				where $\tilde{Z}\sim\mathcal{N}\left(0,E[V_{1}]\right)$,
				$\tilde{G}_{2}:\mathbb{R}^{2}\to\mathbb{R}^{2}\times\mathbb{R}^{2}$ is a
					continuous mapping defined as
					\[
						\tilde{G}_{2}\left(\zeta\right) = E\left[V_{1}\max\left(1 - \tau/(\zeta^{T}V_{1}\zeta),0\right)\right],
					\]
				and $\tilde{\zeta}:\mathbb{R}^{2}\to\mathbb{R}^{2}$ is 	a continuous
				mapping, such that $\tilde{\zeta}(\tilde{Z})$ is the unique solution to
					\[
						\tilde{Z} = -E\left[V_{1}I_{(\tau\leq \zeta^{T}V_{1}\zeta)}\right]\zeta.
					\]
				
				\item When $d\geq 3$,
				\begin{equation*}
				n^{1/2+\frac{3}{2d+2}}R_{n}(\beta_{\ast} )\Rightarrow -2 \tilde{\zeta}(\tilde{Z})^{T} \tilde{Z} - \frac{2}{d+2}\tilde{G}_{3}\left(\tilde{\zeta}(\tilde{Z})\right),
				\end{equation*}
			\end{itemize}
		where $\tilde{Z}\sim\mathcal{N}\left(0,E[V_{1}]\right)$, $\tilde{G}_{2}:\mathbb{R}^{d}\to\mathbb{R}$ is a deterministic continuous
			function defined as
			\[
				\tilde{G}_{2}\left(\zeta\right) = E\left[
				\frac{\pi^{d/2} \gamma f_{X}(X_{1})}{\Gamma\left(d/2+1\right)}
				\left(\zeta^{T}V_{1}\zeta\right)^{d/2+1}
				\right],
			\]
		and $\tilde{\zeta}:\mathbb{R}^{d}\to\mathbb{R}^{d}$ js a  continuous mapping, such that $\tilde{\zeta}(\tilde{Z})$ is the unique solution to
				\[
					\tilde{Z} = - E\left[ V_{1}\frac{\pi^{d/2}\gamma f_{X}\left(X_{1}\right)}{\Gamma\left(d/2+1\right)}
						\left(\zeta^{T}V_{1}\zeta\right)^{d}\right]\zeta.
				\]

\subsection{Proof of Theorem \ref{SoS_theorem_LM}}

\label{Sec-Supp-Proof} In this section, we provide a detailed proof for Theorem \ref{SoS_theorem_LM}. As we discussed before, Theorem
\ref{SoS_theorem_LM} could be treated as a non-trivial corollary of Theorem 3
in \cite{blanchet2016sample} and the proving techniques follow the 6-step
proof for Sample-out-of-Sample (SoS) Theorem, namely Theorem 1 and Theorem 3
in \cite{blanchet2016sample}. 

\begin{proof}
[Proof of Theorem \ref{SoS_theorem_LM}]
\textbf{Step 1.} For $u\in\mathcal{D}_{n}$ and $v\in
\mathcal{X}_{N}$, let us denote $u_{x},u_{y}$ and $v_{x},v_{y}$ to be its
subvectors for the predictor and response. By the definition of RWP function
as in \eqref{Eqn-RWP}, we can write it as a linear program (LP), given as
\begin{align*}
R_{n}\left(  \beta_{\ast}\right)  =\min_{\pi} &  \big\{\sum_{u\in
\mathcal{D}_{n}}\sum_{v\in\mathcal{X}_{N}}\pi\left(  u,v\right)  \left(
\left\Vert u_{x}-v_{x}\right\Vert _{2}^{2}I_{v_{y}=u_{y}}+\infty I_{v_{y}\neq
u_{y}}\right)  \\
&\qquad\qquad \text{ \textbf{s.t. \ }}\pi\in\mathcal{P}\left(  \mathcal{X}%
_{N}\times\mathcal{D}_{n}\right)  ,\\
&  \qquad\qquad\sum_{u\in\mathcal{D}_{n}}\sum_{v\in\mathcal{X}_{N}}\pi\left(
u,v\right)  v_{x}\left(  v_{y}-\beta_{\ast}^{T}v_{x}\right)  =0,\\
& \qquad\qquad\sum
_{v\in\mathcal{X}_{N}}\pi(u,v)=1/n,\forall u\in\mathcal{D}_{n}.\big\}
\end{align*}

For as $n$ large enough the LP is finite and feasible (because $P_{n}$
approaches $P_{\ast}$, and $P_{\ast}$ is feasible). Thus, for $n$ large enough
we can write
\begin{align*}
R_{n}\left(  \beta_{\ast}\right)  =\min_{\pi} &  \big\{\sum_{u\in
\mathcal{D}_{n}}\sum_{v_{x}\in\left\{  X_{i}\right\}  _{i=1}^{N}}\pi\left(
u,v_{x}\right)  \left\Vert u_{x}-v_{x}\right\Vert _{2}^{2}\\
&\qquad\qquad \text{
\ \textbf{s.t. \ \ }}\pi\in\mathcal{P}\left(  \mathcal{X}_{N}\times
\mathcal{D}_{n}\right)  \\
&  \qquad\qquad\sum_{u\in\mathcal{D}_{n}}\sum_{v\in\mathcal{X}_{N}}\pi\left(
u,v\right)  v_{x}\left(  u_{y}-\beta_{\ast}^{T}v_{x}\right)  =0,\\
&\qquad\qquad \sum
_{v\in\mathcal{X}_{N}}\pi(u,v)=1/n,\forall u\in\mathcal{D}_{n}.\big\}
\end{align*}

We can apply strong duality theorem for LP, see
\cite{luenberger1973introduction}, and write the RWP function in dual form:%

\begin{align*}
R_{n}\left(  \beta_{\ast}\right)   &  =\max_{\lambda}\left\{  \frac{1}{n}%
\sum_{i=1}^{n}\min_{j=\overline{1,N}}\left\{  -\lambda^{T}X_{j}\left(
Y_{i}-\beta_{\ast}^{T}X_{j}\right)  +\left\Vert X_{i}-X_{j}\right\Vert
_{2}^{2}\right\}  \right\}  ,\\
&  =\max_{\lambda}\big\{\frac{1}{n}\sum_{i=1}^{n}-\lambda^{T}X_{i}\left(
Y_{i}-\beta_{\ast}^{T}X_{i}\right)  \\
&  +\min_{j=\overline{1,N}}\left\{  \lambda^{T}X_{i}\left(  Y_{i}%
-\beta_{\ast}^{T}X_{j}\right)  -\lambda^{T}X_{j}\left(  Y_{i}-\beta_{\ast}%
^{T}X_{j}\right)  +\left\Vert X_{i}-X_{j}\right\Vert _{2}^{2}\right\}
\big\}.
\end{align*}
This finishes Step 1 as in the 6-step proving technique introduced in Section
3 of \cite{blanchet2016sample}.

\textbf{Step 2 and Step 3}, When $d=1$ and $2$, we consider scaling the RWP
function by $n$ and let define $\zeta=\sqrt{n}\lambda/2$ and denote
$W_{n}=n^{-1/2}\sum_{i=1}^{n}X_{i}e_{i}$, we have the scaled RWP function
becomes,
\begin{align*}
nR_{n}&\left(  \beta_{\ast}\right)  =   \max_{\zeta}\big\{-\zeta^{T}W_{n}\\
&  +\sum_{i=1}^{n}\min_{j=\overline{1,N}}\{-2\frac{\zeta^{T}}{\sqrt{n}}%
X_{j}\left(  Y_{i}-\beta_{\ast}^{T}X_{j}\right)  +2\frac{\zeta^{T}}{\sqrt{n}%
}X_{i}\left(  Y_{i}-\beta_{\ast}^{T}X_{i}\right)  +\left\Vert X_{i}%
-X_{j}\right\Vert _{2}^{2}\}\big\}.
\end{align*}
For each fixed $i$, let us consider the inner minimization problem,
\[
\min_{j=\overline{1,N}}\{-2\frac{\zeta^{T}}{\sqrt{n}}X_{j}\left(  Y_{i}%
-\beta_{\ast}^{T}X_{j}\right)  +2\frac{\zeta^{T}}{\sqrt{n}}X_{i}\left(
Y_{i}-\beta_{\ast}^{T}X_{i}\right)  +\left\Vert X_{i}-X_{j}\right\Vert
_{2}^{2}\}
\]
Similar to Section 3 in \cite{blanchet2016sample}, we would like to solve the
minimization problem by first replacing $X_{j}$ by $a$, which is a free
variable without support constraint in $\mathbb{R}^{d}$, then quantify the
gap. We then obtain a lower bound for the optimization problem via
%\[
%\min_{a}\{-2\frac{\zeta^{T}}{\sqrt{n}}a\left(  Y_{i}-\beta_{\ast}^{T}a\right)
%+2\frac{\zeta^{T}}{\sqrt{n}}X_{i}\left(  Y_{i}-\beta_{\ast}^{T}X_{i}\right)
%+\left\Vert X_{i}-a\right\Vert _{2}^{2}\}.
%\]
\begin{equation}\label{Eqn-Inner-min-lower}
\min_{a}\{-2\frac{\zeta^{T}}{\sqrt{n}}a\left(  Y_{i}-\beta_{\ast}^{T}a\right)
+2\frac{\zeta^{T}}{\sqrt{n}}X_{i}\left(  Y_{i}-\beta_{\ast}^{T}X_{i}\right)
+\left\Vert X_{i}-a\right\Vert _{2}^{2}\}.
\end{equation}
As we can observe in \eqref{Eqn-Inner-min-lower}, the coefficient of second
order of $a$ is of order $O\left(  1/\sqrt{n}\right)  $ for any fixed $\zeta$,
and the coefficients for the last term is always $1$, it is easy to observe
that, as $n$ large enough, \eqref{Eqn-Inner-min-lower} has an optimizer in the interior.

We can solve for the optimizer $a=\bar{a}_{\ast}\left(  X_{i},Y_{i}%
,\zeta\right)  $ of the lower bound in \eqref{Eqn-Inner-min-lower} satisfying
the first order optimality condition as
\begin{align}
\bar{a}_{\ast}\left(  X_{i},Y_{i},\zeta\right)  -X_{i} &  =\left(
e_{i}I-\beta_{\ast}^{T}X_{i}\right)  \frac{\zeta}{\sqrt{n}}%
\label{Eqn-inner-opt}\\
&  +\left(  \beta_{\ast}^{T}\left(  \bar{a}_{\ast}\left(  X_{i},Y_{i}%
,\zeta\right)  -X_{i}\right)  I-\left(  \bar{a}_{\ast}\left(  X_{i}%
,Y_{i},\zeta\right)  -X_{i}\right)  \beta_{\ast}^{T}\right)  \frac{\zeta
}{\sqrt{n}}.\nonumber
\end{align}
Since the optimizer $\bar{a}_{\ast}\left(  X_{i},Y_{i},\zeta\right)  $ is in
the interior, it is easy to notice from \eqref{Eqn-inner-opt} that $\bar
{a}_{\ast}\left(  X_{i},Y_{i},\zeta\right)  -X_{i}=O\left(  \frac{\left\Vert
\zeta\right\Vert _{2}}{\sqrt{n}}\right)  $. Plug in the estimate back into
\eqref{Eqn-inner-opt} obtain
\begin{equation}\label{Eqn-inner-opt-estimation}
	\bar{a}_{\ast}\left(  X_{i},Y_{i},\zeta\right)  =X_{i}+\left(  e_{i}%
	I-\beta_{\ast}^{T}X_{i}\right)  \frac{\zeta}{\sqrt{n}}+O\big(\frac{\left\Vert
	\zeta\right\Vert _{2}^{2}}{n}\big).
\end{equation}
%\[
%\bar{a}_{\ast}\left(  X_{i},Y_{i},\zeta\right)  =X_{i}+\left(  e_{i}%
%I-\beta_{\ast}^{T}X_{i}\right)  \frac{\zeta}{\sqrt{n}}+O\big(\frac{\left\Vert
%\zeta\right\Vert _{2}^{2}}{n}\big).
%\]
Let us define ${a}_{\ast}\left(  X_{i},Y_{i},\zeta\right)  =X_{i}+\left(
e_{i}I-\beta_{\ast}^{T}X_{i}\right)  \frac{\zeta}{\sqrt{n}}$. Using
\eqref{Eqn-inner-opt-estimation}, we have
%\[
%\left\Vert {a}_{\ast}\left(  X_{i},Y_{i},\zeta\right)  -\bar{a}_{\ast}\left(
%X_{i},Y_{i},\zeta\right)  \right\Vert _{2}=O\big(\frac{\left\Vert
%\zeta\right\Vert _{2}^{2}}{n}\big).
%\]
\begin{equation}\label{Eqn-inner-opt-estimation2}
	\left\Vert {a}_{\ast}\left(  X_{i},Y_{i},\zeta\right)  -\bar{a}_{\ast}\left(
	X_{i},Y_{i},\zeta\right)  \right\Vert _{2}=O\big(\frac{\left\Vert
	\zeta\right\Vert _{2}^{2}}{n}\big).
\end{equation}
Then, for the optimal value function of lower bound of the inner optimization
problem, we have:
\begin{align}
&  -2\frac{\zeta^{T}}{\sqrt{n}}\bar{a}_{\ast}\left(  X_{i},Y_{i},\zeta\right)
\left(  Y_{i}-\beta_{\ast}^{T}a\right)  +2\frac{\zeta^{T}}{\sqrt{n}}%
X_{i}\left(  Y_{i}-\beta_{\ast}^{T}X_{i}\right)  +\left\Vert X_{i}-\bar
{a}_{\ast}\left(  X_{i},Y_{i},\zeta\right)  \right\Vert _{2}^{2}\nonumber\\
&  =-2\frac{\zeta^{T}}{\sqrt{n}}{a}_{\ast}\left(  X_{i},Y_{i},\zeta\right)
\left(  Y_{i}-\beta_{\ast}^{T}a\right)  +2\frac{\zeta^{T}}{\sqrt{n}}%
X_{i}\left(  Y_{i}-\beta_{\ast}^{T}X_{i}\right)  \nonumber\\
& \qquad\qquad\qquad+\left\Vert X_{i}-{a}_{\ast
}\left(  X_{i},Y_{i},\zeta\right)  \right\Vert _{2}^{2}+O\big(\frac{\left\Vert
\zeta\right\Vert _{2}^{3}}{n^{3/2}}\big)\nonumber\\
&  =\frac{\zeta^{T}V_{i}\zeta}{n}+O\big(\frac{\left\Vert \zeta\right\Vert
_{2}^{3}}{n^{3/2}}\big).\label{Eqn-lower-bound-est2}%
\end{align}
For the above equation, first equality is due to \eqref{Eqn-inner-opt-estimation2} and the second equality is by the 
estimation of $\bar{a}_{\ast}\left(  X_{i},Y_{i},\zeta\right) $ in 
\eqref{Eqn-inner-opt-estimation}.
%The final equation is due to the estimation of the optimizer (POINT TO
%EQUATION NUMBER HERE).

Then for each fixed $i$, let us define a point process
\[
N_{n}^{(i)}\left(  t,\zeta\right)  =\#\left\{  X_{j}:\left\Vert X_{j}%
-{a}_{\ast}\left(  X_{i},Y_{i},\zeta\right)  \right\Vert _{2}^{2}\leq
t^{2/d}/n^{2/d},X_{j}\neq X_{i}\right\}  .
\]
We denote $T_{i}\left(  n\right)  $ to be the first jump time of $N_{n}%
^{(i)}\left(  t,\zeta\right)  $, i.e.
\[
T_{i}(n)=\inf\left\{  t\geq0:N_{n}^{(i)}\left(  t,\zeta\right)  \geq1\right\}
.
\]
It is easy to observe that, as $n$ goes to infinity, we have
\[
N_{n}^{(i)}\left(  t,\zeta\right)  |X_{i}\Rightarrow Poi\left(  \Lambda
(X_{i},\zeta),t\right)  ,
\]
where $Poi\left(  \Lambda(X_{i},\zeta),t\right)  $ denotes a Poisson point
process with rate
\[
\Lambda(X_{i},\zeta)=\gamma f_{X}\left(  X_{i}+\frac{\zeta}{2\sqrt{\zeta}%
}\right)  \frac{\pi^{d/2}}{\Gamma\left(  d/2+1\right)  }.
\]
Then, the conditional survival function for $T_{i}(n)$, i.e. $P\left(
T_{i}(n)\geq t|X_{i}\right)  $ is
\[
P\left(  T_{i}(n)\geq t|X_{i}\right)  =\exp\left(  -\Lambda\left(  X_{i}%
,\zeta\right)  t\right)  \left(  1+O\left(  1/n^{1/d}\right)  \right)  ,
\]
and we can define $\tau_{i}$ to be the random variable with survival function
being
\[
P\left(  \tau_{i}(n)\geq t|X_{i}\right)  =\exp\left(  -\Lambda\left(
X_{i},\zeta\right)  t\right)  .
\]
We can also integrate the dependence on $X_{i}$ and define $\tau$ satisfying
\[
P\left(  \tau\geq t\right)  =E\left[  \exp\left(  -\Lambda\left(  X_{1}%
,\zeta\right)  t\right)  \right]  .
\]

Therefore,for $d=1$ by the definition of $T_{i}\left(  n\right)  $ and the
estimation in \eqref{Eqn-lower-bound-est2}, we have the scaled RWP function
becomes
\[
nR_{n}\left(  \beta_{\ast}\right)  =\max_{\zeta}\big\{-2\zeta W_{n}-\frac
{1}{n}\sum_{i=1}^{n}\max\big(\zeta^{T}V_{i}\zeta-T_{i}(n)^{2}/n+O\big(\frac
{\left\Vert \zeta\right\Vert _{2}^{3}}{n^{3/2}}\big),0\big)\big\}
\]
The sequence of global optimizers is tight as $n\rightarrow\infty$, because
according to Assumption C, $E(V_{i})$ is assumed to be strictly positive
definite with probability one. In turn, from the previous expression we can
apply Lemma 1 in \cite{blanchet2016sample} and use the fact that the variable
$\zeta$ can be restricted to compact sets for all $n$ sufficiently large. We
are then able to concludee
%\[
%nR_{n}\left(  \beta_{\ast}\right)  =\max_{\zeta}\big\{-2\zeta^{T}%
%W_{n}-E\left[  \max\big(\zeta^{T}V_{i}\zeta-T_{i}(n)^{2}/n,0\big)\right]
%\big\}+o_{p}(1).
%\]

\begin{equation}\label{Eqn-Scaled_RWP-d1}
	nR_{n}\left(  \beta_{\ast}\right)  =\max_{\zeta}\big\{-2\zeta^{T}%
	W_{n}-E\left[  \max\big(\zeta^{T}V_{i}\zeta-T_{i}(n)^{2}/n,0\big)\right]
	\big\}+o_{p}(1).
\end{equation}

When $d=2$, a similar estimation applies as for the case $d=1$. the scaled RWP
function becomes
%\[
%nR_{n}\left(  \beta_{\ast}\right)  =\max_{\zeta}\big\{-2\zeta^{T}%
%W_{n}-E\left[  \max\big(\zeta^{T}V_{i}\zeta-T_{i}(n)^{2},0\big)\right]
%\big\}+o_{p}(1).
%\]
\begin{equation}\label{Eqn-Scaled_RWP-d2}
	nR_{n}\left(  \beta_{\ast}\right)  =\max_{\zeta}\big\{-2\zeta^{T}%
	W_{n}-E\left[  \max\big(\zeta^{T}V_{i}\zeta-T_{i}(n)^{2},0\big)\right]
	\big\}+o_{p}(1).
\end{equation}

For the case when $d\geq3$, let us define $\zeta=\lambda/(2n^{\frac{3}{2d+2}%
})$. We follow a similar estimation procedure as in the cases $d=1,2$. We also
define identical auxiliary Poisson point process, we can write the scaled RWP
function to be
\begin{align}
n^{\frac{1}{2}+\frac{3}{2d+2}}R_{n}\left(  \beta_{\ast}\right)  & =\max_{\zeta
}\big\{  -2\zeta^{T}W_{n}\label{Eqn-Scaled_RWP-d3}\\
&  -n^{\frac{1}{2}+\frac{3}{2+2d}-\frac{2}{d}}E\left[  \max\big(n^{\frac{2}%
{2}-\frac{6}{2d+2}}\zeta^{T}V_{i}\zeta-T_{i}(n)^{3/d},0\big)\right]
\big\}+o_{p}(1).\nonumber
\end{align}
This addresses Step 2 and 3 in the proof. \newline

\textbf{Step 4:} when $d=1$, as $n\rightarrow\infty$, we have the scaled RWP
function given in \eqref{Eqn-Scaled_RWP-d1}. Let us use $G_{1}:\mathbb{R}%
\rightarrow\mathbb{R}$ to denote a deterministic continuous function defined
as
\[
G_{1}\left(  \zeta,n\right)  =E\left[  \max\big(\zeta^{T}V_{i}\zeta
-T_{i}(n)^{2}/n,0\big)\right]  .
\]
By Assumption C, we know $EV_{i}$ is positive, thus $G_{1}$ as a function of
$\zeta$ is strictly convex. Thus the optimizer for the scaled RWP function
could be uniquely characterized via the first order optimality condition,
which is equivalent to
\begin{equation}
\zeta_{n}^{\ast}=-\frac{W_{n}}{E\left[  V_{i}\right]  }+o_{p}(1),\text{ as
}n\rightarrow\infty.\label{Eqn-zeta-est-d1}%
\end{equation}
We plug in \eqref{Eqn-zeta-est-d1} into \eqref{Eqn-Scaled_RWP-d1} and let
$n\rightarrow\infty$. Applying the CLT for $W_{n}$ and the continuous mapping
theorem, we have
\begin{align*}
    nR_{n}\left(  \beta_{\ast}\right)  &=2W_{n}^{2}/E\left[  V_{1}\right]
-G_{1}\left(  -\frac{W_{n}}{E\left[  V_{1}\right]  },n\right)  +o_{p}%
(1)\\
&\Rightarrow\frac{\tilde{Z}^{2}}{E\left[  V_{1}\right]  }=\frac{E\left[
X_{1}^{2}e_{1}^{2}\right]  }{E\left[  \left(  e_{1}-\beta{\ast}X_{1}\right)
^{2}\right]  }\chi_{1}^{2},
\end{align*}
where $W_{n}\Rightarrow\tilde{Z}$ and $\tilde{Z}\sim\mathcal{N}\left(
0,E\left[  \left(  e_{1}-\beta{\ast}X_{1}\right)  ^{2}\right]  \right)  $.

We conclude the stated convergence for $d=1$. \newline

\textbf{Step 5:} when $d=2$, as $n\rightarrow\infty$, we have the scaled RWP
function given in \eqref{Eqn-Scaled_RWP-d2}. Let us use $G_{2}:\mathbb{R\times
N}\rightarrow\mathbb{R}$ to denote a deterministic continuous function defined
as
\[
G_{2}\left(  \zeta,n\right)  =E\left[  \max\big(\zeta^{T}V_{i}\zeta
-T_{i}(n)^{2},0\big)\right]  .
\]
Following the same discussion as in Step 4 for the case $d=1$, we know that
the optimizer $\zeta_{n}^{\ast}$ can be uniquely characterized via first order
optimality condition given as
\[
\label{Eqn-zeta-est-d2}W_{n}=-E\left[  V_{1}I_{(\tau\leq\zeta^{T}V_{1}\zeta
)}\right]  \zeta+o_{p}(1),\text{ as }n\rightarrow\infty.
\]
Since we know that the objective function is strictly convex there exist a
continuous mapping, $\tilde{\zeta}:\mathbb{R}^{2}\rightarrow\mathbb{R}^{2}$,
such that $\tilde{\zeta}(W_{n})$ is the unique solution to
\[
W_{n}=-E\left[  V_{1}I_{(\tau\leq\zeta^{T}V_{1}\zeta)}\right]  \zeta.
\]

Then, we can plug-in the first order optimality condition to the value
function, and the scaled RWP function becomes,
\[
n\mathbb{R}_{n}\left(  \beta_{\ast}\right)  =2\tilde{\zeta}(W_{n})^{T}%
W_{n}-G_{2}\left(  \tilde{\zeta}(W_{n}),n\right)  +o_{p}(1).
\]
Applying Lemma 2 of \cite{blanchet2016sample} we can show that as
$n\rightarrow\infty$,
\[
n\mathbb{R}_{n}\left(  \beta_{\ast}\right)  \Rightarrow2\tilde{\zeta}%
(\tilde{Z})^{T}\tilde{Z}-\tilde{\zeta}\left(  \tilde{Z}\right)  ^{T}\tilde
{G}_{2}\left(  \tilde{\zeta}\left(  \tilde{Z}\right)  \right)  \tilde{\zeta
}\left(  \tilde{Z}\right)
\]
where $\tilde{G}_{2}:\mathbb{R}^{2}\rightarrow\mathbb{R}^{2}\times
\mathbb{R}^{2}$ is a continuous mapping defined as
\[
\tilde{G}_{2}\left(  \zeta\right)  =E\left[  V_{1}\max\left(  1-\tau
/(\zeta^{T}V_{1}\zeta),0\right)  \right]  .
\]

This concludes the claim for $d=2$. \newline

\textbf{Step 6:} when $d=3$, as $n\rightarrow\infty$, we have the scaled RWP
function given in \eqref{Eqn-Scaled_RWP-d3}. Let us write $G_{3}%
:\mathbb{R\times N}\rightarrow\mathbb{R}$ to denote a deterministic continuous
function defined as
\[
G_{3}\left(  \zeta,n\right)  =n^{\frac{1}{2}+\frac{3}{2+2d}-\frac{2}{d}%
}E\left[  \max\big(n^{\frac{2}{2}-\frac{6}{2d+2}}\zeta^{T}V_{i}\zeta
-T_{i}(n)^{3/d},0\big)\right]  .
\]
Same as discussed in Step 4 and 5, the objective function is strictly convex
and the optimizer could be uniquely characterized via first order optimality
condition, i.e.
\[
W_{n}=-E\left[  V_{1}\frac{\pi^{d/2}\gamma f_{X}\left(  X_{1}\right)  }%
{\Gamma\left(  d/2+1\right)  }\left(  \zeta^{T}V_{1}\zeta\right)  ^{d}\right]
\zeta+o_{p}(1),\text{ as }n\rightarrow\infty.
\]
Since we know that the objective function is strictly convex, there exist a
continuous mapping, $\tilde{\zeta}:\mathbb{R}^{d}\rightarrow\mathbb{R}^{d}$,
such that $\tilde{\zeta}(W_{n})$ is the unique solution to
\[
W_{n}=-E\left[  V_{1}\frac{\pi^{d/2}\gamma f_{X}\left(  X_{1}\right)  }%
{\Gamma\left(  d/2+1\right)  }\left(  \zeta^{T}V_{1}\zeta\right)  ^{d}\right]
\zeta.
\]
Let us plug-in the optimality condition and the scaled RWP function becomes
\[
n^{\frac{1}{2}+\frac{3}{2d+2}}R_{n}\left(  \beta_{\ast}\right)  =-2\tilde
{\zeta}(W_{n})^{T}W_{n}-G_{3}\left(  \tilde{\zeta}(W_{n},n)\right)  +o_{p}(1).
\]
As $n\rightarrow\infty$, we can apply Lemma 2 in \cite{blanchet2016sample} to
derive estimation for the RWP function and it leads to
\[
n^{\frac{1}{2}+\frac{3}{2d+2}}R_{n}\left(  \beta_{\ast}\right)  \Rightarrow
-2\tilde{\zeta}(\tilde{Z})^{T}\tilde{Z}-\frac{2}{d+2}\tilde{G}_{3}\left(
\tilde{\zeta}(\tilde{Z})\right)  ,
\]
where $\tilde{G}_{2}:\mathbb{R}^{d}\rightarrow\mathbb{R}$ is a deterministic
continuous function defined as
\[
\tilde{G}_{2}\left(  \zeta\right)  =E\left[  \frac{\pi^{d/2}\gamma f_{X}%
(X_{1})}{\Gamma\left(  d/2+1\right)  }\left(  \zeta^{T}V_{1}\zeta\right)
^{d/2+1}\right]  .
\]
This concludes the case when $d\geq3$ and for Theorem \ref{SoS_theorem_LM}.
\end{proof}

\end{document}